\documentclass[12pt]{article}
\usepackage{times}
\usepackage{graphicx}
\usepackage{color}
\usepackage{multirow}
\usepackage[authoryear]{natbib}
\usepackage{rotating}
\usepackage{bbm}
\usepackage{latexsym}
\usepackage{upgreek}
\usepackage{amsfonts, amsmath, amsthm, amssymb}
\usepackage{ulem}
\usepackage{wasysym}
\usepackage{hyperref}
\usepackage{appendix}
\usepackage{setspace} 
\usepackage{caption}
\usepackage{float}
\usepackage{lineno}
\DeclareMathOperator{\E}{\mathbb{E}}
\newtheorem{definition}{Definition}
\newtheorem{remark}{Remark}
\newcommand{\KL}{\text{KL}} 
\usepackage{etoolbox} 
\makeatletter 
\newcommand*\linenomathpatch{\@ifstar{\linenomathpatch@AMS}{\linenomathpatch@}}
\newcommand*\linenomathpatch@[1]{
  \expandafter\pretocmd\csname #1\endcsname {\linenomathWithnumbers}{}{}
  \expandafter\pretocmd\csname #1*\endcsname{\linenomathWithnumbers}{}{}
  \expandafter\apptocmd\csname end#1\endcsname {\endlinenomath}{}{}
  \expandafter\apptocmd\csname end#1*\endcsname{\endlinenomath}{}{}
}
\newcommand*\linenomathpatch@AMS[1]{
  \expandafter\pretocmd\csname #1\endcsname {\linenomathWithnumbersAMS}{}{}
  \expandafter\pretocmd\csname #1*\endcsname{\linenomathWithnumbersAMS}{}{}
  \expandafter\apptocmd\csname end#1\endcsname {\endlinenomath}{}{}
  \expandafter\apptocmd\csname end#1*\endcsname{\endlinenomath}{}{}
}
\let\linenomathWithnumbersAMS\linenomathWithnumbers
\patchcmd\linenomathWithnumbersAMS{\advance\postdisplaypenalty\linenopenalty}{}{}{}
\makeatother 

\linenomathpatch{equation}
\linenomathpatch*{gather}
\linenomathpatch*{multline}
\linenomathpatch*{align}
\linenomathpatch*{alignat}
\linenomathpatch*{flalign}
\newcommand{\captionfonts}{\normalsize}

\makeatletter  
\long\def\@makecaption#1#2{%
  \vskip\abovecaptionskip
  \sbox\@tempboxa{{\captionfonts #1: #2}}%
  \ifdim \wd\@tempboxa >\hsize
    {\captionfonts #1: #2\par}
  \else
    \hbox to\hsize{\hfil\box\@tempboxa\hfil}%
  \fi
  \vskip\belowcaptionskip}
\makeatother   

\begin{document}
{\LARGE Active inference: demystified and compared} \\

{\bf \large Noor Sajid$^{\displaystyle 1}$},
{\bf \large Philip J. Ball$^{\displaystyle 2}$},
{\bf \large Thomas Parr $^{\displaystyle 1}$}  and 
{ \bf \large Karl J. Friston$^{\displaystyle 1}$}\\
{$^{\displaystyle 1}$Wellcome Centre for Human Neuroimaging, UCL Queen Square Institute of Neurology, London, UK WC1N 3AR.}\\
{$^{\displaystyle 2}$Machine Learning Research Group, Department of Engineering Science, University of Oxford.}\\
\\
\\
\textbf{Correspondence}: Noor Sajid \\
The Wellcome Centre for Human Neuroimaging, \\
UCL Queen Square Institute of Neurology, \\
London, UK WC1N 3AR. \\
+44 (0)20 3448 4362  \\
noor.sajid.18@ucl.ac.uk
\\
\ \\[-2mm]
\thispagestyle{empty}
\markboth{}{}
\ \vspace{-0mm}\\

\begin{center} {\bf Abstract} \end{center}
Active inference is a first principle account of how autonomous agents operate in dynamic, non-stationary environments. This problem is also considered in reinforcement learning, but limited work exists on comparing the two approaches on the same discrete-state environments. In this paper, we provide: 1) an accessible overview of the discrete-state formulation of active inference, highlighting natural behaviors in active inference that are generally engineered in reinforcement learning; 2) an explicit discrete-state comparison between active inference and reinforcement learning on an OpenAI gym baseline. We begin by providing a condensed overview of the active inference literature, in particular viewing the various natural behaviors of active inference agents through the lens of reinforcement learning. We show that by operating in a pure belief-based setting, active inference agents can carry out epistemic exploration --- and account for uncertainty about their environment --- in a Bayes--optimal fashion. Furthermore, we show that the reliance on an explicit reward signal in reinforcement learning is removed in active inference, where reward can simply be treated as another observation we have a preference over; even in the total absence of rewards, agent behaviors are learned through preference learning. We make these properties explicit by showing two scenarios in which active inference agents can infer behaviors in reward-free environments compared to both Q-learning and Bayesian model-based reinforcement learning agents; by placing zero prior preferences over rewards and by learning the prior preferences over the observations corresponding to reward. We conclude by noting that this formalism can be applied to more complex settings; e.g., robotic arm movement, Atari games, etc., if appropriate generative models can be formulated. In short, we aim to demystify the behavior of active inference agents by presenting an accessible discrete state-space and time formulation, and demonstrate these behaviors in a OpenAI gym environment, alongside reinforcement learning agents.


\section{Introduction} Active inference provides a framework (derived from first principles) for solving and understanding the behavior of autonomous agents in situations requiring decision-making under uncertainty \citep{KarlActiveInference2017,KarlDeepTemporal2018}. It uses the free energy principle to describe the properties of random dynamical systems (such as an agent in an environment), and by minimizing expected free energy over time, Bayes--optimal behavior can be obtained for a given environment \citep{dopamine2014,KarlPhysics2019}. More concretely, this optimal behavior is determined by evaluating evidence (i.e., marginal likelihood) under an agent's generative model of  outcomes \citep{KarlLearning2016}. The agent's generative model of the environment is an abstraction, which assumes certain internal (hidden) states give rise to these outcomes. One goal of the agent is to infer what these hidden states are, given a set of outcomes. The generative model also provides a way, through searching and planning, to form beliefs about the future. Thus, the agent can make informed decisions over which sequence of actions (i.e., policies) it is most likely to choose. In active inference, due to its Bayesian formulation, the most likely policies lead to Bayes--optimal outcomes (i.e., those most coherent with prior beliefs). This formulation has two complementary objectives: 1) infer Bayes--optimal behavior, and 2) optimize the generative model based on the agent’s ability to infer which hidden states gave rise to the observed data. Both can be achieved, simultaneously, by minimizing free energy functionals. This free energy formulation gives rise to realistic behaviors, such as natural exploration-exploitation trade-offs, and --- by being fully Bayesian --- is amenable to on-line learning settings, where the environment is non-stationary. This follows from the ability to model uncertainty over contexts \citep{KarlEpistemics2015,TomEpistemics2017}. 

Active inference can also be seen as providing a formal framework for jointly optimizing action and perception \citep{millidge2020relationship}. In the context of machine learning, this is often referred to as planning as inference \citep{Attias2003,Botvinick2012,Baker2014,millidge2020relationship}, and in the case of non-equilibrium physics, it is analogous to self-organization or self-assembly \citep{Crauel1994,Seifert2012,KarlPhysics2019}.

The main contributions of active inference, in contrast to analogous reinforcement learning (RL) frameworks, follow from its commitments to a pure belief-based scheme. Reinforcement learning is a broad term used in different fields. To make meaningful comparisons between active inference and reinforcement learning, we commit to the definition of reinforcement learning in \citep{Sutton1998,sutton2018}: \textit{``reinforcement learning is learning what to do -- how to map situations to actions -- so as to maximize a numerical reward signal"}. RL algorithms, under this definition, can be model-based or model-free. Model-based methods learn a model of environmental dynamics which is used to infer a policy that maximizes long-term reward, while model-free RL estimates a policy that maximizes long-term reward directly from trajectory data. Throughout the paper, RL refers to both model-based and model-free unless stated otherwise. This definition rests on the reward hypothesis, that \textit{``any goal or purpose can be well thought of as maximization of the expected value of the cumulative sum of a received scalar signal (reward)"} \citep{sutton2018}. Here, reward is — by definition — some outcome that reinforces behavior, and hence a circular definition of reward behavior. For example going to the cafe and buying coffee can be explained from two perspectives: 1) a cup of coffee in the morning is intrinsically rewarding and therefore I go out to the cafe to get the coffee and 2) going to the cafe to get coffee means, tautologically, that coffee is rewarding. In short, rewards reinforce behaviors that secure rewards. Traditionally, these (model-free) RL algorithms operate directly on the state of the environment, but (model-based) RL algorithms that operate on beliefs also represent an active area of research \citep{Igl2018}. 

Conversely, in active inference an agent's interaction with the environment is determined by action sequences that minimize expected free energy (and not the expected value of a reward signal). Additionally, unlike in reinforcement learning, the reward signal is not differentiated from other types of sensory outcomes. That is any type of outcome may be more or less preferred. This means that the implicit reward associated with any outcome is a feature of the creature seeing the observation - not the environment they inhabit. This may be different for different agents, or even for the same agent at different points in time. This highlights that the two frameworks have fundamentally different objectives: reward-maximization in reinforcement learning and free energy minimization in active inference. 

In this paper, we reveal circumstances in which behavior might be the same and when it may differ under these two distinct objectives. We show that the main contributions of active inference, in comparison to reinforcement learning, include:
$a)$ reward functions (i.e., prior preferences) are not always necessary because any policy has an epistemic value, even in the absence of prior preferences, $b)$ agents can learn their own reward function and this becomes a way of describing how the agent expects itself to behave —-- as opposed to getting something from the environment, $c)$ a principled account of epistemic exploration and intrinsic motivation as minimizing uncertainty \citep{TomEpistemics2017,Schwartenbeck2019} and $d)$ incorporating uncertainty as a natural part of belief updating \citep{TomEpistemics2017}. Why are these contributions of interest? In standard reinforcement learning, the reward function defines the agent's goal and allows it to learn how to best act within the environment \citep{Sutton1998}. However, defining the reward function is difficult; if it is specific signal from the environment based on action, then is it unanimously good in that environment; e.g., should an agent controlling the thermostat only get positive reward for turning on the heating during winter? Consequently, crafting appropriate reward functions is not easy, and it is possible for agents to learn sub-optimal actions, if the reward function is poorly specified \citep{rewardhacking}. However, active inference bypasses this problem by replacing the traditional reward function, used in reinforcement learning, with prior beliefs about preferred outcomes. This causes the agent to act in a way --- via the beliefs it holds --- such that the observed outcomes match prior preferences. This is useful when we have imprecise or no prior preferences; since active inference endows agents with the ability to learn prior preferences from interacting with the environment itself --- by learning and empirical prior distribution over preferred outcomes. In other words, an agent can learn the kinds of outcomes it can achieve, and these become its prior preferences (in virtue of the fact that these outcomes are achievable, they underwrite the agent’s viability in that environment). This way of defining the reward function (prior preferences) highlights that whether a state is rewarding (or not) is a function of the agent themselves, and not the environment. This reward function conceptualization is distinct from reward functions under reinforcement learning. 

Another challenge within reinforcement learning is balancing the ratio between exploration and exploitation; i.e., what actions should the agent take at any given point in time? Should the agent continue to explore and find more valuable actions or exploit its (current) most valuable action sequence? Many different algorithms have been used to address this; including $\epsilon$-greedy \citep{Vermorel2005,Mnih2013,Mnih2016}, action selection based on action-utility \citep{Sutton1990} and counter-based strategies \citep{Wiering1998,Tijsma2016}, etc. However, even with these exploratory mechanisms in place, most reinforcement learning formulations, call on a temperature hyper-parameter, to weight extrinsic reward (from the environment) against the intrinsic motivation (from the agent). There is no such hyper-parameter in active inference --- although the precision of various priors plays an analogous role --- because the distinction between extrinsic value (i.e., expected reward) and intrinsic value (i.e., intrinsic motivation) is just one way of decomposing expected free energy. In active inference, everything minimizes free energy, including hyper-parameters. Usually, these hyper-parameters transpire to be precision over various beliefs. This allows for a natural trade-off between epistemic exploration and pragmatic behavior. This means that all the required machinery is in play from the start but should only be added to the reinforcement framework (by definition) if it helps maximize expected return. Consequently, state-of-art reinforcement learning approaches can be regarded as a series of refinements to the base algorithm that help resolve problems as they are encountered including the need to marginalize out hyper-parameters instead of defining particular values e.g., \citep{CesaBianchi2017}.

Here, we unpack these properties of active inference, with appropriate ties to the reinforcement learning literature, under the discrete state-space and time formulation; thereby providing a brief overview of the theory. Furthermore, we demonstrate these properties, and points of contact, with reinforcement learning agents on a series of experiments using a modified FrozenLake OpenAI baseline. This is purely an illustration of the conceptual premises; not a demonstration of their implications. Thus, while our simulations of reinforcement learning could have included more complex (i.e., context-aware) aspects –- such as in  \citep{Cao2012,Lloyd2013,Padakandla2019} --— these approaches describe various ways to perform inference, without explicit reference to their impact on behavior. Indeed, as shown by \citep{Donoghue2020}, state-of-the-art approaches, which can be seen as framing reinforcement learning as probabilistic inference, make simplifications for practical reasons. These subtle modeling choices, that trade-off tractability for accuracy, can result in sub-optimal behavior (e.g., failing to account for Q-value uncertainty, and ensuing ‘dithering’ behavior). Active inference avoids this ambiguity by clearly defining how latent variable models are constructed to solve POMDP problems, and how inference should proceed, using gradient descent on expected free energy. This allows the agent to consider the effect of its own actions upon future rewards (i.e., preferred outcomes) when evaluating the expected free energy of all plausible policies (i.e., action trajectories), based upon the anticipated consequences of those policies. However, these are Bellman--optimal for one time-step, but Bayes--optimal (i.e., realize prior beliefs and minimise free energy) for distal time horizons. Additionally, by minimising expected free energy, the agent balances exploration and exploitation resulting in a Bayes-optimal arbitration between the two – which may not be reward-maximizing from an RL perspective. In contrast to RL, active inference accounts for epistemic uncertainty by operating in an explicitly belief-based framework \citep{levine2018}. Additionally, the conceptual approach of active inference means that all the appropriate terms --- relating to intrinsic value of information  --- are in play from the start but should only be added to the reinforcement framework (by definition) if they help maximize long-term reward. Consequently, state-of-art reinforcement learning approaches can be regarded as a series of refinements to the base algorithm that help resolve problems as they are experienced. Therefore, for explicit behavioral comparison, we simulate performance of the two frameworks after removal of the reward signal from the FrozenLake environment i.e., a flat value function. In this setting, there is no motivation for adding any additional information gain term because they cannot be justified in terms of increasing expected value of reward.

The review comprises three sections. The first section considers the discrete state-space and time formulation of active inference, and provides commentary on its derivation, implementation, and connections to reinforcement learning. The second section provides a concrete example of the key components of the generative model and update rules in play, using a modified version of OpenAI's FrozenLake environment. Through these simulations, we compared the performance of three types of agents: active inference, Q-learning \citep{Watkins1992} using $\epsilon$-greedy exploration and Bayesian model-based reinforcement learning using Thompson sampling \citep{Poupart2018a} in stationary and non-stationary environments. We note that whilst all agents are able to perform appropriately in a stationary setting, active inference's ability to carry out online planning allows for Bayes--optimal behavior in the non-stationary environment. The simulations demonstrate that in the absence of a reward signal, the active inference exhibits information seeking behavior (to build a better model of its environment), in contrast to Q-learning agents, but similar to the Bayesian reinforcement-learning agent. We make explicit the conceptual differences in reward function under active inference and reinforcement learning through learning of prior preferences which enable the agent to settle into its niche. We highlight that from the perspective of reinforcement learning, this niche might be counter-intuitive i.e., reward minimising. We conclude with a brief discussion of how this formalism could be applied in (more complex) engineering applications; e.g., robotic arm movement, Atari games, etc., if the appropriate underlying probability distribution or generative model can be formulated.

\section{Active Inference}\label{sec:activeinference}
\subsection*{Motivation}
Active inference describes how (biological or artificial) agents navigate dynamic, non-stationary environments \citep{KarlActiveInference2017,KarlDeepTemporal2018}. It postulates that agents maintain homeostasis by residing in (attracting) states that minimize  surprise \citep{Friston2011,Bogacz2017}.

\begin{linenomath*}
\begin{definition}[Surprise]\label{surprisedefinition} is defined as the negative log probability of an outcome. For this, we introduce a random variable, $o \in O$, that corresponds to a particular outcome received by the agent and $O$ is a finite set of all possible outcomes.
\begin{equation}
    S(o) = -\log P(o)
\end{equation}
where $P$ denotes a probability distribution.
\end{definition}
\end{linenomath*}

In active inference, the agent determines how to minimize surprise by maintaining a generative model of the (partially observable) world. This is necessary because the agent does not have access to a `true' measurement of its current state (i.e., the state of the actual generative process). Instead, it can only perceive itself and the world around via outcomes \citep{KarlActiveInference2017,FristonParr2017}. This allows the problem to be framed as a partially observable Markov decision process (POMDP) \citep{Astrom1965}, where the generative model allows us to make inferences about `true' states given outcomes. In active inference, the agent makes choices based on its beliefs about these states of the world and not based on the ‘value’ of the states \citep{KarlLearning2016}. This distinction is key: in standard model-based reinforcement learning frameworks the agent is interested in optimizing the \textit{value function of the states} \citep{Sutton1998}; i.e., making decisions that maximize expected value. In active inference, we are interested in optimizing a \textit{free energy functional of beliefs about states}; i.e., making decisions that minimize expected free energy. Put even more simply, in reinforcement learning we are interested in residing in high-value states under a reward function, whilst in active inference we wish to reside in states that give rise to outcomes that match our prior preferences (i.e., a target distribution). In one sense, this is a false distinction, as we could interpret a reward function as a log prior preference or vice versa. However, ensuring consistency with a distribution is different to maximizing reward. The latter implies we try to spend all of our time in the most rewarding state, while the former nuances this with an imperative to spend time in states that generate outcomes proportionate to the prior probability associated with those outcomes. However, particular RL algorithms e.g., SMM \citep{lee2019}, also optimize similar objectives of matching state marginal distributions to some target distribution.

From an implementation perspective, this means replacing the traditional reward function used in reinforcement learning with prior beliefs about preferred outcomes. The agent’s prior preferences, $\log P(o)$, are defined only to within an additive constant (i.e., a single negative or positive number). This means that the prior probability of an outcome is a softmax function of utility: $P(o) = \sigma(U(o))$ and therefore depends on relative differences between rewarding \citep{chong2016reconstructing} and unrewarding (surprising) outcomes. Additionally, by incorporating learning - over priors - active inference agent can be equipped with the ability to learn their own preferences over outcomes. Thereby, bypassing the need for an explicit reward signal from the environment and the agent to exhibit self-evidencing behavior (see Section \ref{learningprioroutcomepreferences}).

\subsection*{Variational Free Energy}

\begin{figure}
    \hfill
    \begin{center}
    \includegraphics[width=0.4\linewidth,keepaspectratio]{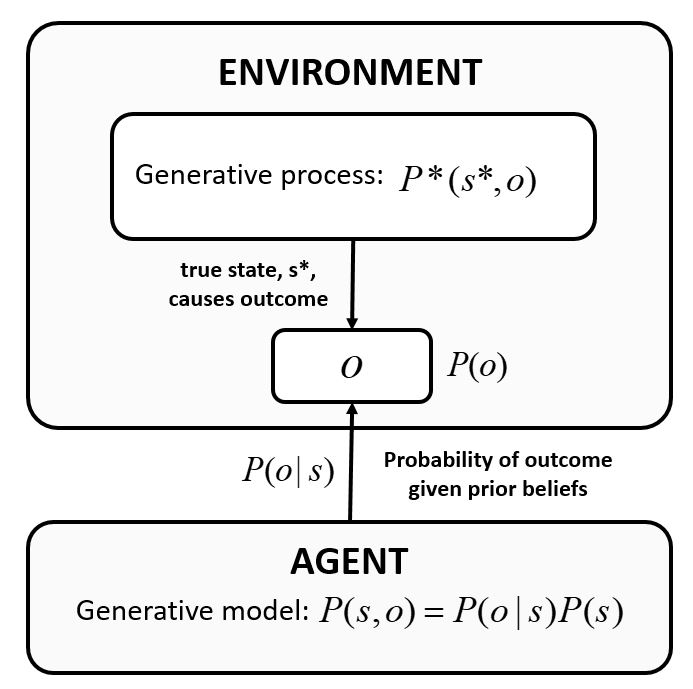}
    \end{center}
    \caption{Graphical representation of the generative process (based on true states, $s*$) in the world and the corresponding (internal) generative model (based on probabilistic beliefs random variables, $s$, that stand in for true states that are hidden) that best explain the outcomes, $o$. This graphic, highlights that the outcomes are shared between the generative process and model.}
    \label{fig:figure1}
\end{figure}

Starting from a simple generative model for outcomes, it is possible to derive a variational free energy formulation, as motivated by Figure~\ref{fig:figure1}; this gives the starting point for the full active inference derivation. First, we introduce the random variable, $s \in S$, to represent a particular hidden state of the world, where $S$ is a finite set of all possible (hidden) states.

The generative model abstraction asserts that the world has a true (hidden) state $s*$, which results in the outcomes $o$ (via the generative process);  $s* \in S$. The agent correspondingly has an internal representation of (or distribution over) $s$, which it infers from $o$ (via its generative model). The hidden state is a combination of features relevant to the agent (e.g., location, color, etc.) and the outcome is the information from the environment (e.g., feedback, velocity, reward, etc.). By the reverse process of mapping from its hidden state to the outcomes (through Bayesian model inversion), the agent can explain the outcomes in terms of how they were caused by hidden states. This is Bayesian model inversion or inference.

\begin{linenomath*}
\begin{definition}[Generative Model] is defined as a partially observable MDP that rests on the following (simplified) joint probability: $P(o,s)$ where $o \in O$ and $s \in S$ as stated previously. The joint probability can be factorized into a likelihood function $P(o|s)$ and prior over internal states $P(s)$ (see supplementary materials for a full specification of the generative model):  
\begin{equation}
    P(o,s) = P(o|s)P(s).
\end{equation}
\end{definition}
\end{linenomath*}

We know that for the agent to minimize its surprise, we need to marginalize over all possible states that could lead to a given outcome. This can be achieved by using the above factorization:
\begin{linenomath*}
\begin{equation}
    P(o) = \sum_{s\in S} P(o|s)P(s) 
\end{equation}
\end{linenomath*}

This is not a trivial task, since the dimensionality of the hidden state (and sequences of actions) space can be extremely large. Instead, we utilize a variational approximation of this quantity, $P(o)$, which is tractable and allows us to estimate quantities of interest.

\begin{definition}[Variational free energy] Variational free energy, $F$, is defined as the upper bound on surprise; definition \ref{surprisedefinition}. It is derived using Jensen's inequality and commonly, known as the (negative) evidence lower bound ($ELBO$) in the variational inference literature \citep{Blei2017}:
\begin{align}
    - \log P(o) &=  - \log \sum_{s \in S} P(o,s) \\
    &\leq - \sum_{s \in S} Q(s) \log \frac{P(o,s)}{Q(s)} \\
    &= \sum_{s \in S} Q(s) \log \frac{Q(s)}{P(o,s)}  
\end{align}
Here, $Q(.)$ is the variational distribution. 

To make the link more concrete, we further manipulate the variational free energy quantity, $F$:
\begin{align}
    F &= \sum_{s \in S} Q(s) \log \frac{Q(s)}{P(o,s)}\\
      &= \sum_{s  \in S} Q(s) \log \frac{Q(s)}{P(s|o)P(o)}\\
      &= \sum_{s  \in S} Q(s) \left( \log \frac{Q(s)}{P(s|o)} - \log P(o)\right)\\
      &= D_\KL[Q(s)||P(s|o)] - \log P(o)
\end{align}
By rearranging the last equation, the connection between surprise and variational free energy is made explicit:
\begin{align}
    - \log P(o) &= F - D_\KL[Q(s)||P(s|o)]
\end{align}
Additionally, we can express variational free energy as a function of these posterior beliefs in many forms:
\begin{align}
    F &= \underbrace{D_\KL[Q(s|\pi)||P(s|o,\pi)]}_{\substack{\text{evidence bound}}} - \underbrace{\log P(o)}_{\substack{\text{log evidence}}} \label{eq:com} \\
     &= \underbrace{D_\KL[Q(s|\pi)||P(s|\pi)]}_\text{complexity} - \underbrace{\mathbb{E}_{s\sim Q(s)}[\log P(o|s)}_\text{accuracy}] \label{eq:kl}
\end{align}
\end{definition}

Since KL divergences cannot be less than zero, from Equation \ref{eq:com} we see that free energy is minimized when the approximate posterior becomes the true posterior. In that instance, the free energy would simply be the negative log evidence for the generative model \citep{Beal2003}. This shows that minimizing free energy is equivalent to maximizing (generative) model evidence. In other words, it is minimizing the complexity of accurate explanations for observed outcomes, as seen in Equation \ref{eq:kl}. Note that we have conditioned the probabilities in Equation \ref{eq:com} and \ref{eq:kl} on policies, $\pi$. These policies can be regarded as hypotheses about how to act that, as we will see below, pertain to probabilistic transitions among hidden states. For the moment, the introduction of policies simply means that the variational free energy above can be evaluated for any given action sequence.

\subsection*{Expected Free Energy}

Variational free energy gives us a way to perceive the environment (i.e., determine $s$ from $o$), and addresses one part of the problem; namely, making inferences about the world (i.e., the `inference' in active inference). However, the `active' part of the formulation is still lacking; we have not accounted for the fact that the agent can take actions. To motivate this, we note that we would like to minimize not only our variational free energy $F$ calculated from past and present observations, but also our expected free energy $G$, which depends upon anticipated observations in the future. Minimization of expected free energy allows the agent to influence the future by taking actions in the present, which are selected from policies. We will first consider the definition of a policy, and later determine how to evaluate their likelihoods from the generative model, which ultimately leads to the action selected by the agent.

\begin{definition}[Policy]\label{policydefinition} is defined as a sequence of actions at time, $\uptau$, that enable an agent to transition between hidden states; $\uptau \in [1,2,...,T]$, where $T$ is the total number of time-steps in a given experiment under the generative model.
For this, we introduce two random variables: 1)
$u_{\uptau} \in U$ to represent a particular action at time $\uptau$ where $U$ is a finite set of all possible actions, and 2) $\pi \in \Pi$. to represent a particular policy, where $\Pi$ is a finite set of allowable policies i.e., sequences of actions in the sense of sequential policy optimization; e.g., \citep{Alagoz2010}.
\begin{align}
    \pi=\{u_1,u_2, ...,u_{\uptau}\}
\end{align}
up to a given time horizon, $\uptau$. The explicit link between policy and action is
\begin{align}
    u_\uptau = \pi(\uptau)
\end{align}
\end{definition}

\begin{remark}[Connections to state-action policies] 
From definition \ref{policydefinition}, in active inference a policy is simply a sequence of choices for actions through time i.e., a sequential policy. This contrasts with state-action policies in reinforcement learning, where  $\pi_{RL}$ denotes a mapping of states to actions; e.g. \citep{Bellman1952,Sutton1990}
\begin{align*}
   \pi_{RL}(u,s) = P(u\mid s)
\end{align*}
The two policy types -- sequential, $\pi$ and state-action $\pi_{RL}$ -- are equivalent, under a POMDP formulation, when $\uptau = 1$ i.e., $\pi=\{u_1\}$ \citep{KarlLearning2016}. Note that the Bayes--optimal policy is selected from these policies. For the remainder of the paper, a policy ($\pi$) refers to sequential policy. 
\end{remark}

To derive the expected free energy, we first extend the variational free energy definition to be dependent on time ($\uptau$) and policy ($\pi$) (and present its matrix formulation: Equation~\ref{eq:matrix}):
\begin{align}
    F(\uptau,\pi) &= \sum_{s^\pi_\uptau} Q(s_\uptau|\pi)Q(s_{\uptau-1}|\pi) \log \frac{Q(s_\uptau|\pi)}{P(o_\uptau,s_\uptau|s_{\uptau-1},\pi)}\label{eq:gref} \\
    &= \E_{Q(s_{\uptau-1}|\pi)}\big[D_\KL[Q(s_\uptau|\pi)||P(s_\uptau|s_{\uptau-1},\pi)]\big] - \E_{Q(s_\uptau|\pi)}\big[\ln P(o_\uptau|s_\uptau)\big] \\
    &=s^\pi_\uptau\cdot\big(\log s^\pi_\uptau - \log \boldsymbol{B}^\pi_{\uptau-1}s^\pi_{\uptau-1} - \log \boldsymbol{A}\cdot o_\uptau\big)\label{eq:matrix} 
\end{align}
Here $s^\pi_\uptau$ is the expected state conditioned on each policy; $\boldsymbol{B}^\pi_{\uptau}$ is the transition probability for hidden states, contingent upon pursuing a given policy, at a particular time; $\boldsymbol{A}$ is the expected likelihood matrix mapping from hidden states to outcomes and $o_\uptau$ represents the outcomes. These are simply vectors ($s^\pi_\uptau$) or matrices ($\boldsymbol{B}^\pi_{\uptau}$ and $\boldsymbol{A}$) specifying a probability for each alternative state or outcome. For the matrices each column corresponds to a different value to the variable we condition upon (here, hidden states), while rows give the probability of each hidden state at the next time or the outcome at the current time step, respectively.  Now having developed this functional dependency on time, we simply take an expectation with respect to the posterior distribution of outcomes from our generative model, $P(o_\uptau|s_\uptau)$.

\begin{definition}[Expected free energy]\label{Gdefinition} is defined as a free energy functional of future trajectories, $G$. It effectively evaluates evidence for plausible policies based on outcomes that have yet to be observed \citep{ParrG2018}. Heuristically, we can obtain $G$ from Equation~\ref{eq:gref} by making two moves. The first is to include beliefs about future outcomes in the expectation; i.e., supplementing the expectation under the approximate posterior with the likelihood, resulting in a predictive distribution given by $P(o_\uptau|s_\uptau)Q(s_\uptau|\pi)$. The second is to (implicitly or explicitly) condition the joint probabilities of states and observations in the generative model upon some desired state of affairs (C), as opposed to a specific policy. These two moves ensure (1) we can evaluate this quantity before the observations are obtained and (2) minimization of G encourages policies whose result is consistent with C \footnote{There are other ways in which we could consider constructing free energy functionals to deal with outcomes that have yet to be observed. While some of the alternatives are plausible from a theoretical perspective, they tend to dispense with aspects of observed behavior that we seek to capture – and therefore do not apply to the kinds of system we are interested in here. For example, Millidge et~al \citep{millidge2020} propose an alternative that subtracts the conditional entropy of the likelihood (i.e., ambiguity) from the expected free energy shown here. This leads to agents who are less ambiguity averse, and do not seek information. As active inference deals with curious agents, we retain this ambiguity in the expected free energy functional.}.

\begin{align}
    G(\uptau,\pi)  &= \sum_{s_\uptau,o_\tau} P(o_\uptau|s_\uptau)  Q(s_\uptau|\pi)Q(s_{\uptau-1}|\pi) \log \frac{Q(s_\uptau|\pi)}{P(o_\uptau,s_\uptau|s_{\uptau-1},C)} \label{eq:g} \\
    &= \E_{\tilde{Q}} \left[\log(Q(s_\uptau|\pi) - \log(P(o_\uptau,s_\uptau|s_{\uptau-1},C))\right]\\
    &= \E_{\tilde{Q}} \left[\log(Q(s_\uptau|\pi) - \log(P(s_\uptau|o_\uptau,s_{\uptau-1})) - \log(P(o_\uptau|C))\right]\\
    &\geq \underbrace{\E_{\tilde{Q}} \left[ \log(Q(s_\uptau|\pi) - \log(Q(s_\uptau|o_\uptau, s_{\uptau-1},\pi))\right]}_{\text{-ve mutual information}}  - \underbrace{\E_{\tilde{Q}}\left[\log(P(o_\uptau|C)) \right]}_{\text{expected log evidence}}\label{eq:v1}\\
    &= \underbrace{\E_{\tilde{Q}} \left[ \log(Q(o_\uptau|\pi) - \log(Q(o_\uptau|s_\uptau,s_{\uptau-1},\pi))\right]}_\text{-ve epistemic value}  - \underbrace{\E_{\tilde{Q}}\left[\log(P(o_\uptau|C)) \right]}_\text{extrinsic value}\label{eq:v2}\\
    &= \underbrace{D_\KL[Q(o_\uptau|\pi)||P(o_\uptau|C)]}_\text{expected cost}+ \underbrace{E_{Q(s_\uptau|s_{\uptau-1},\pi)}\left[H[P(o_\uptau|s_\uptau)] \right]}_\text{expected ambiguity}\label{eq:v3}\\
    &= o^\pi_\uptau \cdot (o^\pi_\uptau - \boldsymbol{C}_\uptau) + s^\pi_\uptau\cdot\boldsymbol{H}\label{eq:matG}
\end{align}
\end{definition}
where the following notation is used: $\tilde{Q} = P(o_\uptau|s_\uptau)Q(s_\uptau|\pi)$; $Q(o_\uptau|s_\uptau,\pi) = P(o_\uptau|s_\uptau)$; $\boldsymbol{C}_\uptau = \log P(o_\uptau|C)$ is the logarithm of prior preference over outcomes, $ o_\uptau $ is the vector of posterior predictive outcomes (i.e., $ \boldsymbol{A}s^\pi_\uptau $) and $\boldsymbol{H}=-diag\big(\E_Q[\boldsymbol{A}_{i,j}].\E_Q[\boldsymbol{A}]\big)$ is the vector encoding the ambiguity over outcomes for each hidden state.

When minimizing expected free energy, we can regard Equation \ref{eq:v2} as capturing the imperative to maximize the amount of information gained, from observing the environment, about the hidden state (i.e., maximizing epistemic value), whilst maximizing expected value – as scored by the (log) preferences (i.e., extrinsic value). This entails a clear trade-off: the former (epistemic) component promotes curious behavior, with exploration encouraged as the agent seeks out salient states to minimize uncertainty about the environment, and the latter (pragmatic) component encourages exploitative behavior, through leveraging knowledge that enables policies to reach preferred outcomes. In other words, the expected free energy formulation enables active inference to treat exploration and exploitation as two different ways of tackling the same problem: minimizing uncertainty. 

This natural curiosity can be contrasted with handcrafted exploration in reinforcement learning schemes, where curiosity is replaced by random action selection \citep{Mnih2013} or through the use of ad hoc novelty bonuses, which are appended to the reward function \citep{Deepak2017}. Information theoretic approaches have also been explored in a reinforcement learning context e.g., \citep{Still2012,Mohamed2015,blau2019bayesian}. Some of these approaches leverage beliefs about latent states \citep{blau2019bayesian,sekar2020planning}. For example, Blau et al, \citep{blau2019bayesian} is a model-free algorithm that implicitly accounts for beliefs over the latent states. Additionally, Seker et al. \citep{sekar2020planning} is very close in its treatment of latent states to active inference but leverages an ensemble of belief states to inform epistemic exploration, rather than a true Bayesian posterior. Additionally, curiosity as formulated under active inference can emerge in reinforcement learning under POMDP formulations. Example algorithms may incorporate inductive biases \citep{Igl2018} and uncertainty over state transitions or outcomes \citep{Ross2008,kolter2009,Schulze2020}. The key aspect of these belief-POMDP schemes is that they deal with belief states (i.e., a space of probability distributions over hidden states). This is crucial for exploration and minimizing uncertainty, because uncertainty is an attribute of a belief about hidden states, not the hidden states per se.

Normatively speaking, active inference dispenses with the Bellman optimality principle and replaces it with a (variational) principle of least action -- please see \citep{KF2012SAM} for further discussion. However, recent Bayesian RL schemes have used variational principles e.g. maintaining latent over the MDP \citep{Schulze2020} or the explicit beliefs \citep{igl2019generalization}. While in many of these settings, the distance between the two schools of thought may seem to be closing, a fundamental distinction—-that has yet to be bridged—-is the situation in which there are no rewards or, in active inference, when prior preferences are uninformative. In a scheme motivated by reward maximization, no meaningful behavior can be generated in this setting. This is not a criticism of such schemes, but a statement of their scope and the problems they are designed to solve. In contrast, the intrinsic value of seeking information—-regardless of its potential to evince reward—-in active inference means that in the absence of any rewarding outcomes, agents are still driven by a curiosity that helps them build a better model of their world. Furthermore, active inference agents can learn their priors over observations, and will exhibit ambiguity minimizing behavior in order to fulfill these prior expectations \citep{KarlLearning2016}. In short, they can learn epistemic habits in the absence of extrinsic rewards. 


Equation \ref{eq:v3} offers an alternative perspective on the same objective; i.e., an agent wishes to minimize the ambiguity and the degree to which outcomes (under a given policy) deviate from prior preferences $P(o_\uptau|C)$. Thus, ambiguity is the expectation of the conditional entropy --- or uncertainty about outcomes --- under the current policy. Low entropy suggests that outcomes are salient and uniquely informative about hidden states (e.g., visual cues in a well-lit environment --- as opposed to the dark). In addition, the agent would like to pursue policy dependent outcomes ($Q(o_\uptau|\pi)$) that resemble its preferred outcomes ($P(o_\uptau|C)$). This is achieved when the KL divergence between predicted and preferred outcomes (i.e. expected cost) is minimized by a particular policy. Furthermore, prior beliefs about future outcomes equip the agent with goal-directed behavior (i.e. towards states they expect to occupy and frequent).

It is now also possible to specify priors over policies using the expected free energy. Policies, a priori, minimize the expected free energy term, $G$ \citep{KarlActiveInference2017}. 
This has sometimes been framed in terms of a heuristic reductio ad absurdum argument that if selected policies realize prior beliefs and minimise free energy, then the only tenable prior beliefs are policies that will minimise free energy \citep{ParrG2018}. If this were not true, then an active inference agent would not have prior belief that it selects policies that minimise expected free energy and it would infer (and pursue) policies that were not free energy minimising. As such, it would not be an active inference (i.e., free energy minimising) agent, which is a contradiction. This leads to the agent’s prior belief that it will select policies that minimise the free energy expected under that policy. There are some subtleties to this argument that leave some room for maneuver. Specifically, this does not tell us how to construct an expected free energy functional -– but this is typically chosen to be consistent with Definition \ref{Gdefinition}. This choice ensures both exploratory and exploitative behavior and is therefore sufficiently flexible to deal with the kind of problem we are interested in for this paper. 

This can be realized by expressing the probability of any policy with a softmax function (i.e., normalized exponential) of expected free energy:
\begin{align}
    P(\pi) = \sigma[-\beta^{-1} \cdot G(\pi)]\label{pG}
\end{align}
where $\sigma$ denotes a softmax function and $\beta$ is a temperature parameter.

This illustrates the `self-evidencing' behavior of active inference. Action sequences (policies) that result in lower expected free energy are more likely. Intuitively this makes sense; since all notions of how to act in the world (i.e., exploration, exploitation) are wrapped up in the expected free energy $G$, policy selection simply becomes a matter of determining (through search) the set of actions which get us closest to this goal (i.e., the attracting set defined by prior preferences $P(o|C)$).

Note the similarities to Dyna-style/planning model-based reinforcement learning \citep{Sutton1990}: hypothetical roll-outs are used to model the consequences of each policy. However, the actual controller in active inference is derived through an approach similar to model predictive control \citep{Camacho2007}, where a search is performed over possible action sequences at each time-step.

Now that we have priors over policies, we can incorporate these into the generative model, and into the free energy. This gives:
\begin{align}
    F = \E_{Q(\pi)}[F(\pi)] + D_\KL[Q(\pi)||P(\pi)]
\end{align}
Here, the free energy of the model conditioned upon the model plays the role of a negative log marginal likelihood, giving this the form of an accuracy and complexity term in the space of beliefs about policies. Often, the temperature parameter $ \beta $ is also equipped with priors (normally using a gamma distribution) and posteriors, which add an additional complexity term to the free energy.

\subsection*{Optimizing Free Energy}\label{subsec:optimize}

From this free energy formulation, we can optimize expectations about hidden states, policies, and precision through inference, and optimize model parameters (likelihood, transition states) through learning (via a learning rate: $\eta$). This optimization requires finding sufficient statistics of posterior beliefs that minimize variational free energy \citep{FristonParr2017}. Under variational Bayes, this would mean iterating the appropriate formulations (for inference and learning) until convergence. Under the active inference scheme, we calculate the solution by using a gradient descent (with a default step size, $\zeta$, of 4) on free energy $F$, which allows us to optimize both action-selection and inference simultaneously, using a mean-field approximation \citep{Beck2012,Parr2019}. The gradients of the negative free energy with respect to states and precisions, respectively, are
\begin{align}
    \varepsilon^\pi_\uptau &= (\log \boldsymbol{A}\cdot o_\uptau + \log \boldsymbol{B}^\pi_{\uptau-1}s^\pi_{\uptau-1} + 
    \log \boldsymbol{B}^\pi_{\uptau}\cdot s^\pi_{\uptau+1}) - \log s^\pi_\uptau \label{up2}\\
    \varepsilon^\gamma &= (\beta - \beta_\uptau) + (\pi - \pi_0) \cdot G \label{up1}
\end{align}
where $\beta_\uptau = \beta + (\pi - \pi_0).G$;  $\beta=\frac{1}{\gamma}$ encodes posterior beliefs about (inverse) precision (i.e., temperature); $\pi$ represents the policies specifying action sequences and $\pi_0 = \sigma(-\gamma.G)$. $G$ in this equation is a vector whose elements are the expected free energies for each policy under consideration.

This involves converting the discrete updates, defined in Equation \ref{up2} and \ref{up1}, into dynamics for inference that minimize state and precision prediction errors: $\varepsilon^\pi_\uptau = -\partial_{s}F$ and $ \varepsilon^\gamma = -\partial_{\gamma}F$ . These prediction errors are free energy gradients. Gradient flows then produce posterior expectations that minimize free energy to provide Bayesian estimates of hidden variables. This particular optimization scheme means expectations about hidden variables are updated over several time scales. During each outcome or trial, beliefs about each policy is evaluated based upon prior beliefs about future outcomes - which get in through the expected free energy. This is determined by updating posterior beliefs about hidden states (i.e., state estimation under each policy, $Q(s|\pi)$) on a fast time scale, while posterior beliefs find new extrema (i.e., as new outcomes are sampled, $P(s|o,\pi)$). These posterior beliefs are then used to compute the posterior predictive probabilities of future outcomes, $Q(o|\pi)$, which themselves contribute to the expected free energy, and through this the priors over policies. This introduces an unusual feature of the online inference schemes under active inference rarely seen in Bayesian accounts - the priors over policies change with new outcomes. This is a key distinction between active inference and standard Bayesian RL perspectives, and can only occur when the best policies to engage in are functionals of beliefs which may be updated. 

Using this kind of belief updating, we can calculate the posterior beliefs about each policy; namely, a softmax function based on expected free energy and the likelihood of past observations under that policy (approximated with $ F(\pi)$), extending the definition of the prior covered in Equation \ref{pG}. The softmax function is a generalized sigmoid for vector input, and can, in a neurobiological setting, be regarded as a firing rate function of neuronal depolarization \citep{KarlDeepTemporal2018}. Having optimized posterior beliefs about policies, they are used to form a Bayesian model average of the next outcome (i.e., under these beliefs about what I will do next, which observations would I expect on average), which is realized through action selected to conform to this. Practically, action selection is often achieved more simply by sampling from the distribution over actions at a given time implied by posterior beliefs about policies.

In active inference, the scope and depth of the policy search is exhaustive, in the sense that any policy entertained by the agent is encoded explicitly, and any hidden state over the sequence of actions entailed by policy are continuously updated. However, in practice, this can be computationally expensive; therefore, a policy is no longer evaluated if its log evidence is $\zeta$ (default $20$) times less likely than the (current) most plausible policy. This, $\zeta$, can be treated as an adjustable hyper-parameter. Additionally, at the end of each sequence of outcomes, the expected parameters are updated to allow for learning across trials. This is like Monte-Carlo reinforcement learning, where model parameters are updated at the end of each trial.
Lastly, temporal discounting emerges naturally from active inference, where the generative model determines the nature of discounting (based on parameter capturing precision), with predictions in the distant future being less precise and thus discounted in a Bayes--optimal fashion \citep{KarlActiveInference2017}. Practically, this involves the inclusion of a hyper-parameter $\beta = \gamma^{-1}$ that can be regarded as the (inverse precision or temperature) of posterior beliefs over policies. This is marginalized out during belief updating; enabling the exploration-exploitation trade-off to be inferred. This highlights the flexibility of active inference; in the sense it can be applied to any generative model. It is unclear how similar parameterizations could be adopted in conventional reinforcement learning, where optimizing temperature parameters typically involves a grid search over value. Having said this, sophisticated formulations of reinforcement learning, e.g., soft actor-critic \citep{Haarnoja2018,CesaBianchi2017} have addressed this issue.
 
The above discussion suggests that, from a generic generative model, we can derive Bayesian updates that clarify how perception, policy selection and actions shape beliefs about hidden states and subsequent outcomes in a dynamic (non-stationary) environment. This formulation can be extended to capture a more representative generative process by defining a hierarchical (deep temporal) generative model as described in \citep{KarlActiveInference2017,FristonParr2017,TomEpistemics2017}, continuous state spaces models \citep{Buckley2017,ParrFriston2019,ueltzhoffer2018deep} or mixed models with both discrete and continuous states as described in \citep{FristonParr2017,ParrFriston2018}. In the case of a continuous formulation, the generative model state-space can be defined in terms of generalized coordinates of motion (i.e., the coefficients of a Taylor series expansion in time - as opposed to a series of discrete time-steps), which generally have a non-linear mapping to the observed outcomes. Additionally, future work looks to evaluate how these formulations (agents) may interact with each other to emulate multi-agent exchanges. 

The implicit variational updates presented here have previously been used to simulate a wide range of neuronal processing (using a gradient descent on variational free energy): ranging from single cell responses (including place-cell activity) \citep{KarlActiveInference2017}, midbrain dopamine activity \citep{dopamine2014}, to evoked potentials, including those associated with mismatch negative (MMN) paradigms \citep{KarlActiveInference2017}. Additionally, there has been some evidence implicating these variational inferences with neuromodulatory systems: action selection (dopaminergic), attention and expected uncertainty (cholinergic) and volatility and unexpected uncertainty (noradrenergic) \citep{TomEpistemics2017,ParrG2018}. Please see \citep{KarlActiveInference2017,ParrG2018,lance2020}, for a detailed overview.

In what follows, we provide a simple worked example to show precisely the behaviors that emerge --- naturally --- under active inference.

\section{Simulations}
\label{sec:1}
This section considers inference using simulations of a modified version of OpenAI gym's FrozenLake environment: for simplicity, we have chosen this paradigm (note that more complex simulations have been explored in the literature; e.g., behavioral economics trust games \citep{Moutoussis2014,Schwartenbeck2015b}, narrative construction and reading \citep{KarlDeepTemporal2018}, saccadic searches and scene construction \citep{Mirza2016}, Atari games \citep{OpenAI2018}, etc). In closely related work, Cullen, et~al \citep{OpenAI2018} demonstrated that active inference agents perform better in another OpenAI gym environment Doom, compared to reward-maximizing agents. Their reward-maximizing agents are active inference agents without the epistemic value term ($G$) and can therefore be considered distinct from standard reinforcement learning agents. 

We first describe the environment set-up and then simulate how an agent learns to navigate the lake to successfully reach the goal. The simulations involve searching for the reward (i.e., Frisbee) in a $3\times3$ frozen lake and avoid falling in a hole. The purpose of these simulations is to provide an accessible overview —-- and accompanying code (see Software Note) --— of the conceptual (and practical) differences between active inference and standard reinforcement learning.

\subsection*{Set-up}
\begin{figure}
    \begin{center}
    \includegraphics[width=0.94\linewidth,keepaspectratio]{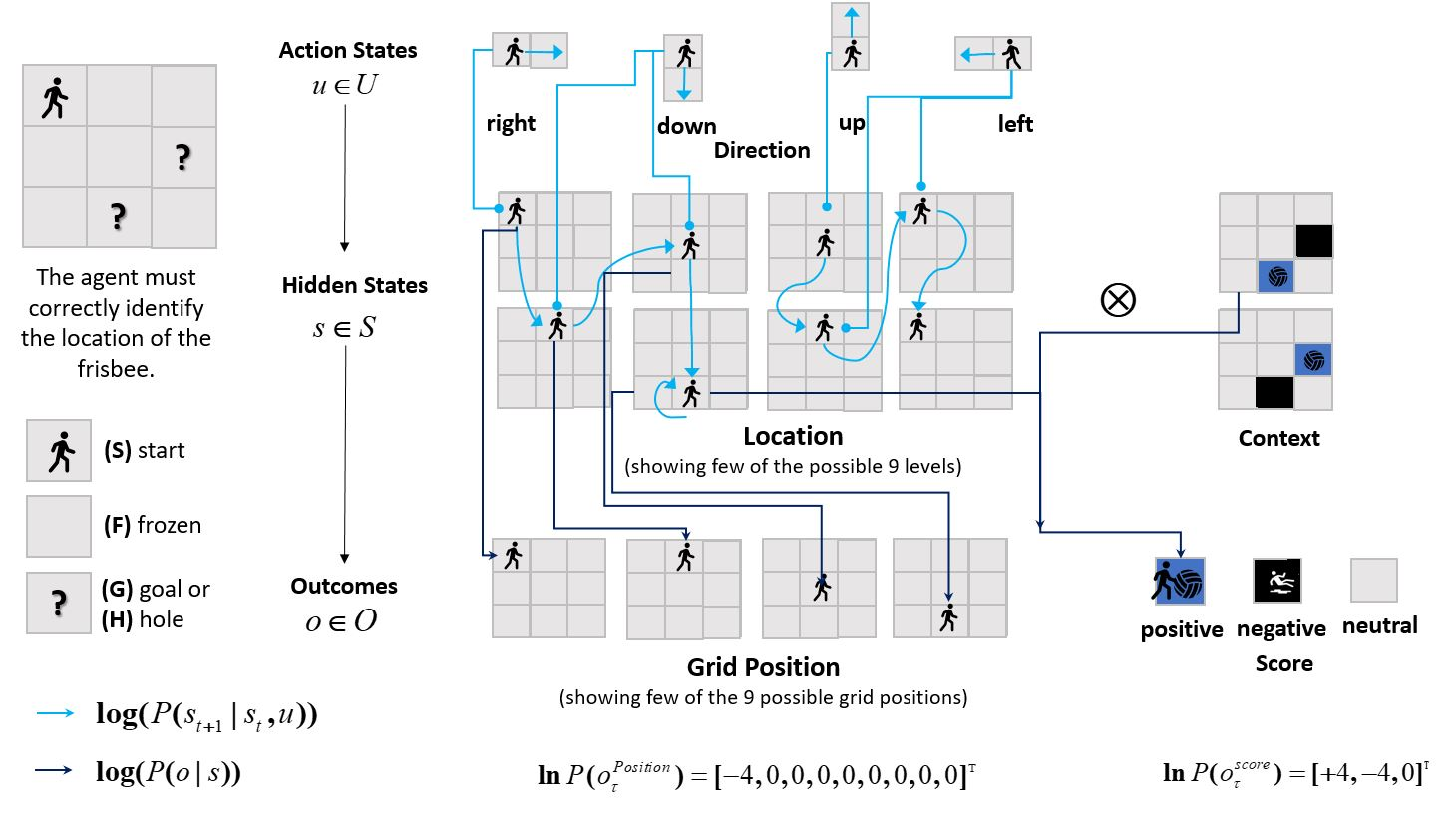}
    \end{center}
    \vspace{-1cm}
    \caption{{Graphical representation of the active inference generative model: The model contains four action states that encode direction of movement: right, down, up and left. They control the ability to transition between hidden state \textit{location} factors (one of the nine locations going from $1 \to 9$ - only a few states are shown). Each action navigates the agent to a different location (a select few are shown): e.g., if the agent starts in position $1$ and chooses to turn right, then it will end up in state $2$ at the next time--step. However, if the agent started in location $5$, and goes up, it would end up in location $2$ instead. Note that both $8$ and $6$ are absorbing states: only $8$ is denoted as such by the circular arrow. Additionally, if an agent makes an improbable move; i.e., tries to go left from location $1$, it will remain in location $1$ (as shown). The hidden states have a Kronecker tensor product ($\otimes $) form with two factors: location and context (one of the two goal locations). The context cannot be changed by the agent and corresponds to the associated Frisbee location: $8$ if context $1$ or $6$ if context $2$. Note that in context $1$, the hole location is $6$. From each of the two hidden state factors (location and context) an outcome is generated. The agent observes two types of outcomes at each time point: its grid position and score. Categorical parameters, that define the generative model, $\boldsymbol{A}$ (likelihood - $P(o|s)$): have an identity mapping between hidden state location and outcome grid position with some uncertainty; e.g. if I have beliefs that I am in position $6$, then I will observe myself in position $6$, irrespective of context. The \textit{score} likelihood, given the hidden states, is determined by the context; i.e., for context $1$, positive score received at location $8$, and negative or nothing elsewhere. $\ln P(o)$ corresponds to prior preference: the agent expects to find positive score and not remain at the starting location.}}
    \label{fig:figure2}
\end{figure}

The frozen lake has a grid-like structure with four different patches: starting point (S), frozen surface (F), hole (H) and goal (G) where the Frisbee is located. All patches, except for (H), are safe. The agent starts each episode at (S); position $1$. From there, to reach the Frisbee location, the agent needs to take a series of actions; e.g. left, right, down or up. The agent is allowed to continue moving around the frozen lake, with multiple revisits to the same positions, but each episode ends when either (H) or (G) is visited. (G) and (H) can be located in one of two locations: position $8$ and $6$ or $6$ and $8$ respectively. The objective is to reach (G), the Frisbee location, ideally in as few steps as possible, whilst avoiding the hole (H). If the agent is able to reach the Frisbee without falling in the hole, it receives a score of $100$ at the end of trial. This scoring metric is framework agnostic and allows us to compare active inference to reinforcement learning methods. However, it is important to note that maximizing reward is not the definition of Bayes--optimal behavior for an active inference agent, where information gain is also of value. This will become important later. Finally, we limit the maximum number of time steps (i.e., the horizon) to 15.

\subsection*{Active inference agents}
For this paradigm, we define the generative model for the active inference agent as follows (Figure~\ref{fig:figure2}): four action states that encode direction of movement (left, right, down and up), $18$ hidden states ($9$ locations factorized by $2$ contexts) and outcome modalities include grid position ($9$) and score ($3$). The action states control the transitions between the hidden state location factors e.g. when at location $4$, the agent can move to location $5$ (right), $7$ (down), $1$ (up) or stay at $4$ (left). The hidden state factor, \textit{location}, elucidates the agents’ beliefs about its location in the frozen lake. The context hidden state factor elucidates the agent’s beliefs about the location of (G) and (H): if context is $1$, then (G) location is $8$ and (H) location is $6$. The outcomes correspond to the following: being at any of the $9$ possible grid positions and receiving $3$ types of potential reward (positive, negative or neutral). Positive reward is received if the agent correctly navigates to the (G) location, negative if to the (H) location and neutral otherwise (F, S). 

We define the likelihood $P(o|s)$ as follows: an identity mapping between hidden state location and outcome grid position; e.g., if I have beliefs that I am located in position $6$, then I will observe myself in position $6$, irrespective of context. However, the likelihood for score, given the hidden states, is determined by the context; i.e. if the context is $1$ ($2$) then positive score will be received at location $8$ ($6$), and negative or nothing elsewhere. The action-specific transition probabilities $P(s_{t-1}|s_t, u)$ encode allowable moves, except for the sixth and eight locations, which are absorbing latent states that the agent cannot leave. We define the agent as having precise beliefs about the contingencies (i.e., large prior concentration parameters $= 100$). The utility of the outcomes, $\boldsymbol{C}$, is defined by $\ln P(o)$ : $4$ and $-4$ $nats$ for rewarding and unrewarding outcome: this can be regarded as a replacement for writing out an explicit reward function. This means, that the agent expects to be rewarded $e^{8}$ times more, at (G) than (H). Notice that rewards and losses are specified in terms of $nats$ or natural units, because we have stipulated reward in terms of the natural logarithms of some outcome. The prior beliefs about the initial state were initialized: location state ($D = 1$) for the first location and zero otherwise, with uniform beliefs for context state. We equip the agent with deep policies: these are potential permutations of action trajectories e.g., ('Left, 'Left', 'Right') or ('Down', 'Right', 'Up'). Practically, policies (action sequences) are removed if the relative posterior probability is of $1/128$ or less than the most likely policy. After each episode, the posteriors about the current state are carried forward as priors for the next episode. By framing the paradigm in this way we treat solving the POMDP as a \textit{planning as inference} problem; in order to act appropriately the agent needs to correctly update internal beliefs about the current context.

Having specified the state-space and contingencies, we can solve the belief updating Equations~\ref{up2} and \ref{up1} to simulate appropriate behavior. Pseudo-code for the belief updating and action selection for this particular type of discrete state-space and time formulation is presented in supplementary materials. To provide a baseline for purely exploratory behavior, we also simulated a ‘null’ active inference agent, who had no prior preferences (i.e., was insensitive to the reward).

\subsection*{Reinforcement learning agents}

We compared the active inference agents' performance against two reinforcement learning algorithms: Q-Learning using $\epsilon$-greedy exploration \citep{Watkins1989,Sutton1998} and Bayesian model-based reinforcement learning using standard Thompson sampling \citep{Poupart2018a,Ghavamzadeh2015}. Thompson sampling is an appropriate procedure here, because it entails the optimization of dual objectives; reward maximization and information gain. This is achieved by having a distribution over a particular function, that is parameterized by a prior distribution over it.

We evaluate two permutations of the Q-learning algorithm, an agent with fixed exploration ($\epsilon = 0.1$) and an agent with decaying exploration ($\epsilon = 1$ decaying to $0$); the pseudo-code is presented in supplementary materials. For both Q-learning agents, we specify the learning rate as $0.5$ and discount factor as $0.99$. 

The Bayesian RL approach is a standard Dyna-style \citep{Sutton1990} approach, where we train Q-learning agents in a belief-based internal model (planning), which accounts for uncertainty over both the transition model and reward function (i.e., separate prior distribution over both functions); the pseudo-code is presented in supplementary materials. The transition model, encodes the probability for the next state, given the current state and action. These transition probability distributions are the same as the active inference generative model above: high probability for intended move and extremely low probability for an implausible move. The reward function, encodes the uncertainty about the reward location (an implicit contextual understanding about the environment). The likelihoods, for the transition model and reward function, are modeled via two separate Bernoulli distributions; with Beta distributions as the conjugate prior over their parameters. The Beta distribution pseudo-counts --- for the reward and transition model --- are initialized as 1. The posterior for the reward and transition model distribution are evaluated by updating the prior ($Beta(\alpha, \beta)$). Thus, by treating them as pseudo-counts, the evidence for intended move (likely reward location), $x$, is added to $\alpha$ and an implausible move (unlikely reward location), $y$, is added to $\beta$: posterior is $Beta(\alpha+x, \beta+y)$. The discount factor is specified as $0.9$.

The Bayesian RL agent is a planning--based RL algorithm that parameterizes the transition and reward model using two separate Bernoulli distribution with a Beta prior. At each episode, we sample k (=50), $\theta$, from each of the Beta distributions. Using the sampled priors, we define k MDPs and solve them using value iteration. This simulation gives us the Q-value function which is averaged out to get the optimal Q-value. The optimal Q-value function is used to determine the next action and move to the next state. This procedure continues till the agent reaches the goal or falls down the hole. The process of solving the k MDP to determine the next action is similar to hypothetical roll-outs in other planning-based algorithms. 

Note that more sophisticated reinforcement learning schemes may have been more apt for solving this task; e.g., \citep{Daw2006,Fuhs2007,Sam2010,Daw2011,Gershman2017}. However, our aim was to compare standard formulations of active inference and reinforcement learning. The rationale for this will become clearer when we compare the behavioral performance in the absence of reward --- no motivation for adding any additional heuristics because they cannot be justified in terms of increasing expected value of reward. Thus, the adopted RL agents are suited for our purpose.

\subsection*{Learning to navigate the frozen lake}
We evaluate how well the different agents are able to navigate the frozen lake in both stationary and non-stationary environments, as described below. Each of the environments were simulated for $200$ trials with $500$ episodes for the five agents: Q-learning ($\epsilon = 0.1$), Q-learning ($\epsilon= 1$ decaying to $0$), Bayesian model-based reinforcement learning, Active Inference (Figure~\ref{fig:figure2}) and Active Inference (null model; without any prior outcome preferences i.e. $\log P(o|C) = 0$ for all outcomes). To aid intuition, the flattening of the prior preferences in the active inference model is equivalent to reclassifying reward as just another state or observation in a reinforcement learning scheme. While an agent would still be 'told' whether or not it had encountered a rewarding stimulus, this would have no impact on the value function. As noted above, this is not an exact equivalence, as there is a philosophical distinction between making a change to the environment (in the reinforcement learning setting) and to the agent (in the active inference setting).

\begin{table}[!htbp]
\centering
\resizebox{\columnwidth}{!}{%
\begin{tabular}{*4c}
\hline
{} &  {} & \multicolumn{2}{c}{Average Score [95\% CI]
}\\
Algorithm   & Belief-Based  & Deterministic Env. & Stochastic Env.\\
\hline
\hline
Q-Learning ($\epsilon=0.1$)  &  N & $97.79$ $[97.41,98.16]$ & $66.08$ $[63.28,68.88]$\\
Q-Learning ($\epsilon=1$ decaying to 0)   &  N & $80.44$ $[78.96,81.93]$  & $65.13$ $[62.57,67.68]$\\
Bayesian RL   &  Y  &  $99.76$ $[99.45, 100.00]$  & $64.39$ $[60.33, 68.44]$\\
Active Inference  &  Y  &  $99.88$ $[99.64, 100.00]$   & $98.90$ $[98.00, 99.79]$\\
Active Inference (null model)  &  Y  &  $50.03$ $[49.70, 50.35]$  & $50.22$ $[49.89, 50.22]$\\
\hline
\end{tabular}%
}
    \caption{Average reward (and $95\%$ confidence interval) for each agent, across both deterministic and stochastic environments. The results are calculated from the $200$ trials across $500$ episodes.}
    \label{tab:results1}
\end{table}

\subsubsection*{Stationary environment}
For this set-up, the goal (G) exists at position $6$ and hole (H) location at $8$ for the entire experiment. We then evaluate the agent performance online, and make no distinction between offline and online behavior modes. This is to better simulate exploration and exploitation in the real world, where we use the same policy to gather training data and act; indeed it is this exact paradigm which is one of the major motivators for active inference. The average score (Table~\ref{tab:results1}) for all agents, except the null model specification of Active inference model, was considerably high at $>80$, showing that all frameworks were able to solve the MDP.

The low score for the null (active inference) model reflects the lack of prior preferences for the type of outcomes the agent would like to observe i.e., it does not differentiate between any of the different patches (S, F, G \& H) in the frozen lake. The null model does not seek out the goal state (i.e., reward), which it does not prefer over other states. Instead, it either falls in the hole or reaches the goal, with equal probability. From the perspective of information seeking, this is a very sensible policy, as the same amount of information can be obtained either by finding the hole (which tells us where the goal is) or by finding the goal (which tells us where the hole is). As such, there is nothing to disambiguate between the two. Over subsequent exposures to the environment, given that it is stationary, this agent will be left with little uncertainty to resolve, as it will know everything at the start of a trial based upon past experience. Over time, this will lead to a loss of purposeful behavior, resembling what we might expect from a reinforcement learning agent in the absence of any environmental rewards (even in the presence of uncertainty).

\begin{figure}
\centering
  \centering
  \includegraphics[width=0.9\linewidth]{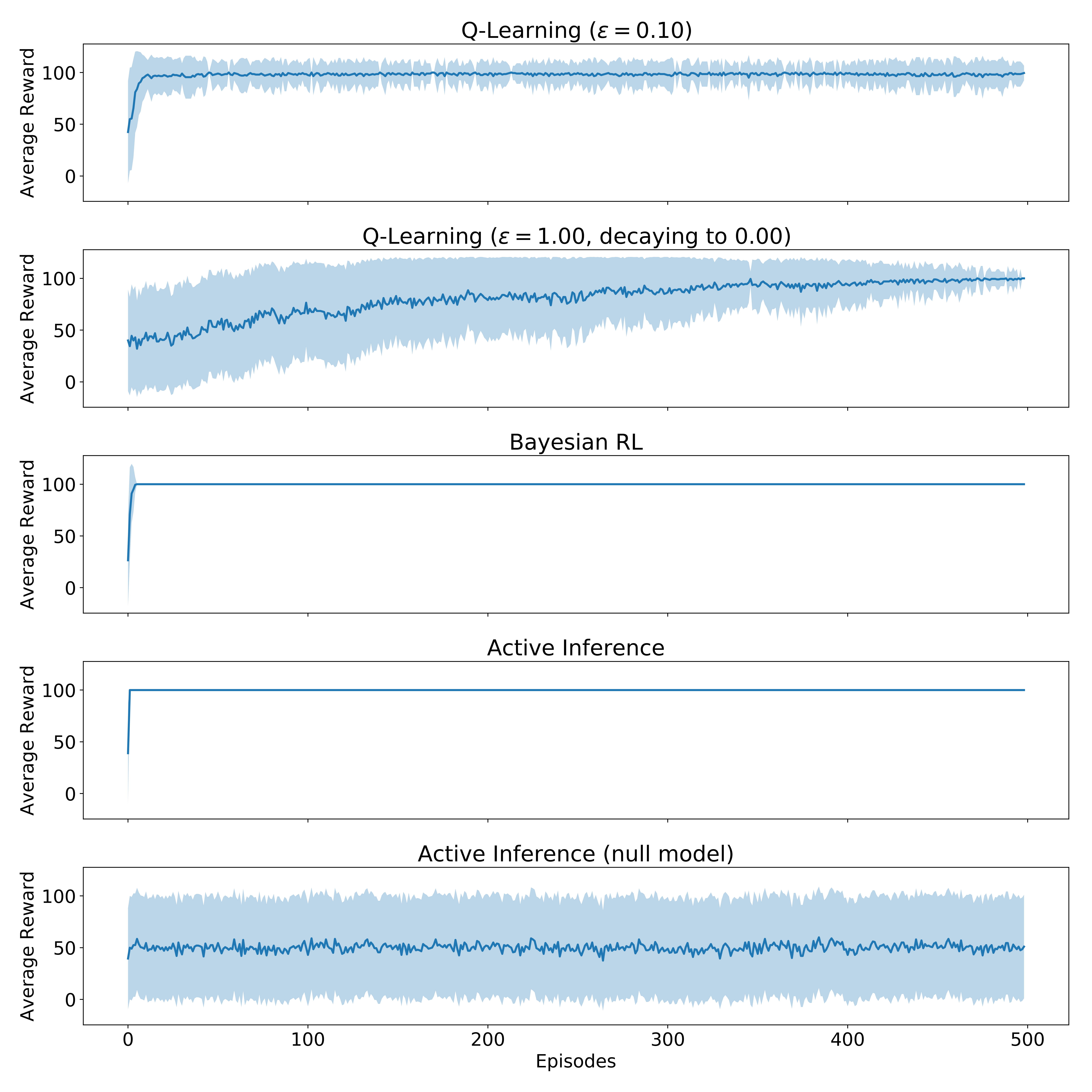}
  \caption{Learning curve for deterministic environment. The x-axis denotes the episode number and y-axis the average (online) reward. The results presented are calculated from $200$ trials.}
  \label{fig:sub1}
\end{figure}

The learning curve, as shown in (Figure~\ref{fig:sub1}), highlights that the active inference and Bayesian model-based reinforcement learning agent learn reward-maximizing behavior (and resolve uncertainty about reward location) in a short amount of time ($<10$ episodes). They are able to maintain this for the remaining trials. This is reflected by the tight confidence intervals around the average reward for both agents. In contrast, Q-learning ($\epsilon = 0.1$), whilst also quickly learning appropriate state-action pairing, has slightly larger confidence intervals for the average reward due to the $10\%$ of selecting a random action.

\subsubsection*{Non-stationary environment}
We introduce non-stationarity into the environment; the location of the (G) and (H) are flipped after a certain number of episodes. Initially (G) is located at position $6$ and (H) at position $8$, and then we swap (G) and (H) at the following time steps: $21, 121, 141, 251, 451$. This means after episode $451$, (G) remains at position $8$ until the end of the simulation. These changes in the reward location test how quickly the agent can re-learn the correct (G) location. The average score for all agents is presented in Table~\ref{tab:results1}.

\begin{figure}
  \centering
  \includegraphics[width=0.9\linewidth]{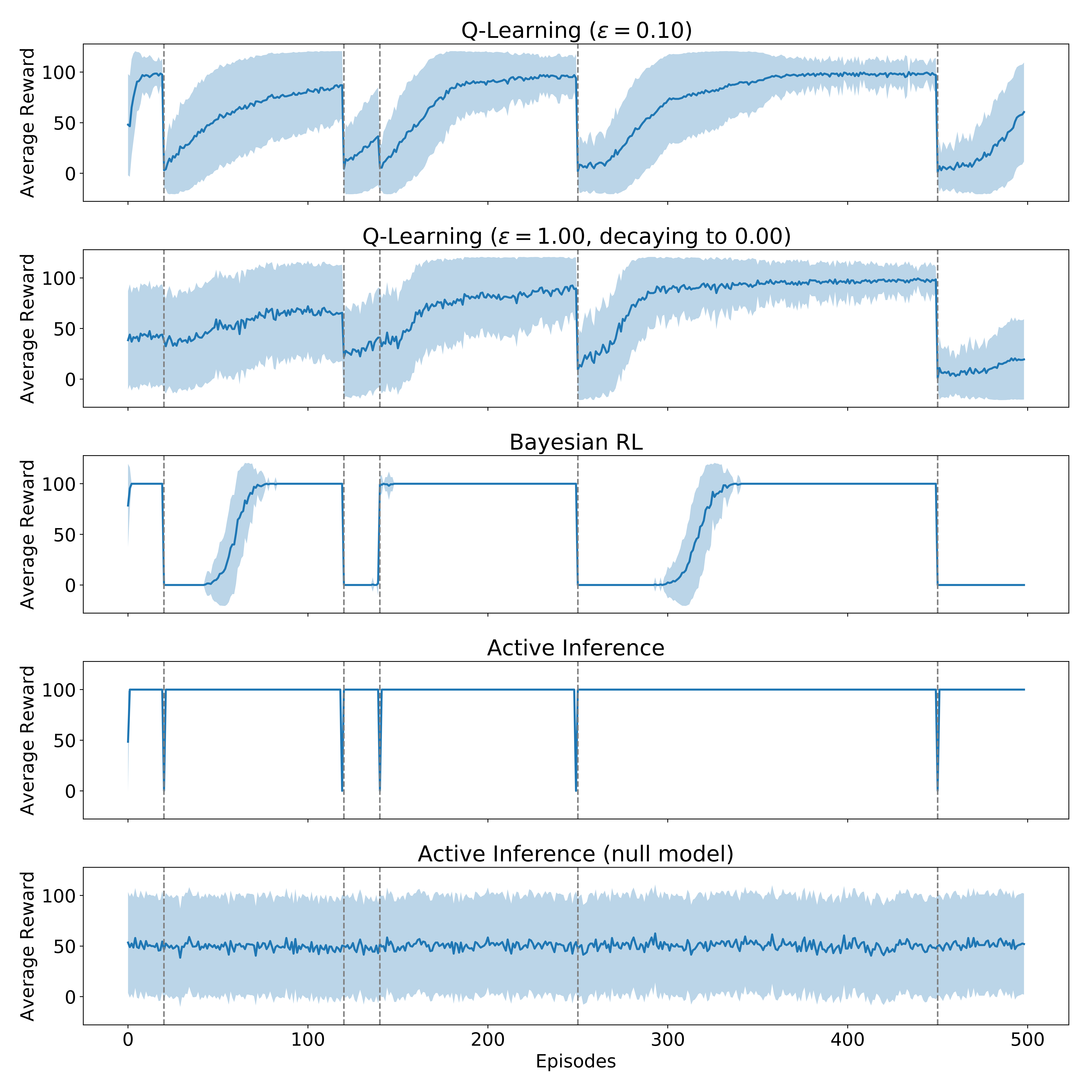}
  \caption{Learning curve for stochastic environment. The x-axis denotes the episode number and y-axis the average (online) reward. The results presented are calculated from $200$ trials. The dotted gray lines represent the change in (G) (and (H)) location.}
  \label{fig:sub2}
\end{figure}

As in the stationary set-up, all agents are initially uncertain about the reward location. This is quickly resolved, and by episode $20$, active inference, Bayesian RL and Q-learning ($\epsilon=0.1$) exhibit appropriate behavior for solving the task. The null (active inference) model and Q-learning ($\epsilon=1$ decaying to $0.00$) exhibit fairly poor performance (consistent with stationary). However, at episode $21$, the performance for all agents drops to $0$ due to the change in reward location --- except for the agent with no preferences, who persists with achieving the reward half of the time. For the reinforcement learning (Q-learning and Bayesian RL) agents, this drop in performance persists for the next $\sim40$ episodes. This is because by treating this as a `learning' problem, the agent has to do the following: 1) reversal learning of its previous understanding of the reward location and 2) re-learn the current reward location. In contrast, by treating this as a \textit{planning as inference} problem, the active inference agent is able to quickly recover performance after a single episode, as the generative model takes into account the context switch. In other words, the agent simply infers that a switch has happened, and acts accordingly. This quick performance recovery is persistent for all changes in reward location across the $500$ episodes (Figure~\ref{fig:sub1}). However, for Bayesian RL the ability to adapt its behavior to the changing goal locations continues to prove difficult; each time a greater number of episodes are required to reverse the learning of the prior distribution over the reward function due to the accumulation of pseudo-counts. This contrasts with Q-learning ($\epsilon=0.1$), which adapts fairly quickly to these fluctuating reward locations, because it needs to only update the appropriate state and action Q-values. 
 
Therefore, for non-stationary environments active inference offers an attractive, natural adaptation mechanism --- for training artificial agents --- due to its Bayesian model updating properties. This is in contrast to standard reinforcement learning, where issues of environmental non-stationarity are not accommodated properly, as shown through the above simulations. They can be dealt with using techniques that involve the inclusion of inductive biases; e.g., importance sampling of experiences in multi-agent environments \citep{JF2017} or using meta-learning to adapt gradient-update approaches more quickly \citep{AS2018}. Lastly, we acknowledge that the simulations presented limit the comparison between active inference and standard (i.e., naive) reinforcement learning schemes. We could have introduced further complexity (i.e., additional distributions to be learnt) within the Bayesian reinforcement learning model: for example, explicit beliefs about latent contexts instead of implicit context encoding, via the reward location \citep{Sam2010,Gershman2015,Rakelly2019}. To evaluate the comparison of more complex reinforcement learning agents, to active inference, in non-stationary environments remains an outstanding research question. We appreciate that with additional design choices the Bayesian reinforcement learning agent may exhibit similar behavioral performance to the active inference agent. 

\subsection*{Comparing prior preferences and rewards}
In reinforcement learning, goals are defined through reward functions i.e., explicit scalar signal from the environment. In contrast, in active inference, goals are defined through the agents prior preferences over outcomes. We now illustrate the link between these definitions of goal-directed behavior by presenting experiments that show the effect of reward shaping \citep{Ng2003} in the FrozenLake stationary environment (Table~\ref{tab:rs_reward}).

We apply the following shaping: modifying the reward for reaching the goal (G), modifying the reward for falling down the hole (F), and modifying the reward for any state that isn't a goal (H) (this can be considered a `living cost'). In order to convert the shaped reward into prior preferences, we manipulate the prior preferences such that their relative weighting matches that introduced through the reward shaping e.g., reward of $-100$ is equivalent to prior preferences of $-\log(5)$, etc.

As our experiments show, when we define a prior preference through a reward function, the behaviors of the belief-based policies (i.e., Bayesian RL and Active Inference) are nearly identical, and learn to solve the environment as soon as a positive reward is defined for the goal. On the other hand, the non-probabilistic Q-learning approach appears more sensitive to reward shaping, with living costs causing greedier behavior (i.e., taking fewer steps per episode). A possible explanation for this is that the construction of the generative models for both Bayesian RL and Active Inference clearly define that the location of the goal/hole is in either states 6 or 8, hence Bayes--optimal behavior (i.e., getting to the goal in as few steps as possible) can be learned even in the absence of negative rewards/preferences over certain states. All that is required is some notion of where the goal state might exist, hence the ability to learn Bayes--optimal policies by only specifying the goal location (see the last row of Table~\ref{tab:rs_reward}).

Another interesting behavior is when there is an absence of preferences/rewards (i.e., first row of Table~\ref{tab:rs_reward}). The Q-learning approach learns a deterministic circular policy with little exploration despite the $\epsilon$ term since it does not update its parameters due to the lack of reward signal. The belief-based approaches on the other hand maintain exploration throughout --- represented by the average score ranging between $~40-45$ --- as their probabilistic models remain uniform over the beliefs of which transitions produce preferred outcomes. This suggests that by having an objective function that is optimizing a dual objective, the agent exhibits some sort information gain i.e., exploring the world is intrinsically motivated because it helps the agent build a better model of the world. We will see more purposeful exploration, under active inference, in the next section.

Finally, we observe that all 3 approaches learn the same circular behavior when only a negative preference or reward is specified (i.e., second row of Table~\ref{tab:rs_reward}). This is because all the approaches learn to avoid the hole state (H), but since there is no notion of goal-seeking behavior, do not learn to go to the goal state. Interestingly, in the case of the belief-based approaches (Bayesian RL and Active Inference), since the generative model defines the presence of hole states in either state 6 or 8, and since it receives no preference for goal states, the generative model assigns non-zero probability with the hole state being in either state 6 or 8. As a result policies derived from these generative models learn to avoid both states, therefore only terminating when the time limit is reached.

Through this brief study, we have illustrated an implicit equivalence between Bayesian model-based reinforcement learning and active inference. This equivalence rests on treating prior preferences as a reward function. In other words, by expressing an arbitrary reward function as a potential function (i.e., a log probability over future outcomes), reward functions can be absorbed into expected free energy. This means one can elicit identical behaviors from reinforcement learning and active inference. Indeed, if one removes uncertainty --- in the form of epistemic value --- we are left with pragmatic value; namely, expected future reward. This shows that reinforcement learning can be regarded as a limiting or special case of model-based approaches in general --- or active inference in particular. However, the FrozenLake environment is by no means representative of all discrete environments, and this merits further research. It is important to note that behavioral equivalences are a result purely of the environmental set-up and the accompanying reward signal, e.g., changing the FrozenLake environment for a maze with noisy attractor state, and no reward, might reveal additional behavioral differences between Bayesian model-based reinforcement learning and active inference. 

\begin{table}[!htbp]
\centering
\begin{tabular}{*6c}
\hline
\multicolumn{3}{c}{Rewards} & \multicolumn{3}{c}{Average Score (Average Number of Moves)} \\
(G) & (H) & (F) & Q-Learning* ($\epsilon = 0.1$) & Bayesian RL & Active Inference\\
\hline
\hline
$0.00$ & $0.00$ & $0.00$ & $0.00$ $(15.00)$ & $39.94$ $(9.17)$ & $44.00$ $(8.67)$ \\
$0.00$ & $-100$ & $0.00$ & $0.00$ $(15.00)$ & $0.00$ $(15.00)$ & $0.00$ $(15.00)$ \\
$100$ & $-100$ & $0.00$  & $95.56$ $(3.53)$ & $99.77$ $(3.02)$ & $99.52$ $(3.03)$ \\ 
$100$ & $0.00$ & $-10.0$ & $96.00$ $(3.48)$ & $99.89$ $(3.00)$ & $99.47$ $(3.00)$\\
$100$ & $-100$ & $-10.0$ & $96.47$ $(3.42)$ & $99.79$ $(3.01)$ & $99.58$ $(3.00)$ \\
$100$ & $0.00$ & $0.00$ & $95.32$ $(3.58)$ & $99.74$ $(3.00)$ & $99.50$ $(3.07)$ \\
\hline
\end{tabular}
    \caption{Reward shaping: average score and number of moves across $100$ episodes for $100$ agents. *Note that for this experiment we evaluate under $\epsilon = 0.0$, i.e., on-policy.}
    \label{tab:rs_reward}
\end{table}

\subsubsection*{Learning prior outcome preferences}\label{learningprioroutcomepreferences}
In some settings, explicitly defining prior outcome preferences might be challenging due to time dependent preferences, an inability to disambiguate between different types of outcomes, or simply lack of domain knowledge. In those instances, the appropriate distribution of prior outcome preferences can be learned via the agent’s interaction with the environment. This difficulty extends to reinforcement learning, where defining a reward function may not be possible, and in its vanilla formulation, reinforcement learning offers no natural way to learn behaviors in the absence of a reward function (see the first row of Table~\ref{tab:rs_reward}).

In order to demonstrate the ability of active inference to select policies in the absence of pre-specified prior preferences, we allow both the likelihood distribution ($\log P(o|s)$) and outcome preferences ($\log P(o|C)$) to be learned. This allows us to make explicit that whether a state is rewarding or not is determined by the agent learning its prior preferences and it is not a specific signal from the environment. For this, the generative model is extended to include prior beliefs about the parameters of these two distributions (a prior over priors in the case of ($\log P(o|C)$), which are learned through belief updates \citep{KarlActiveInference2017}. The natural choice for the conjugate prior for both distributions is a Dirichlet distribution, given that the probability distributions are specified as a categorical distribution. This means that the probability can be represented simply in terms of Dirichlet concentration parameters. We define the Dirichlet distribution (for both likelihood and prior preferences) as completely flat (initialized as $5$ for likelihood and $1$ for prior preferences for all possible options). This is in contrast to row one of Table~\ref{tab:rs_reward}, where we specify flat prior preferences, but the agent is not equipped with (Dirichlet) hyper-priors that enable the agent to learn about the kind of outcomes it prefers.

\begin{figure}
  \centering
  \includegraphics[width=\linewidth]{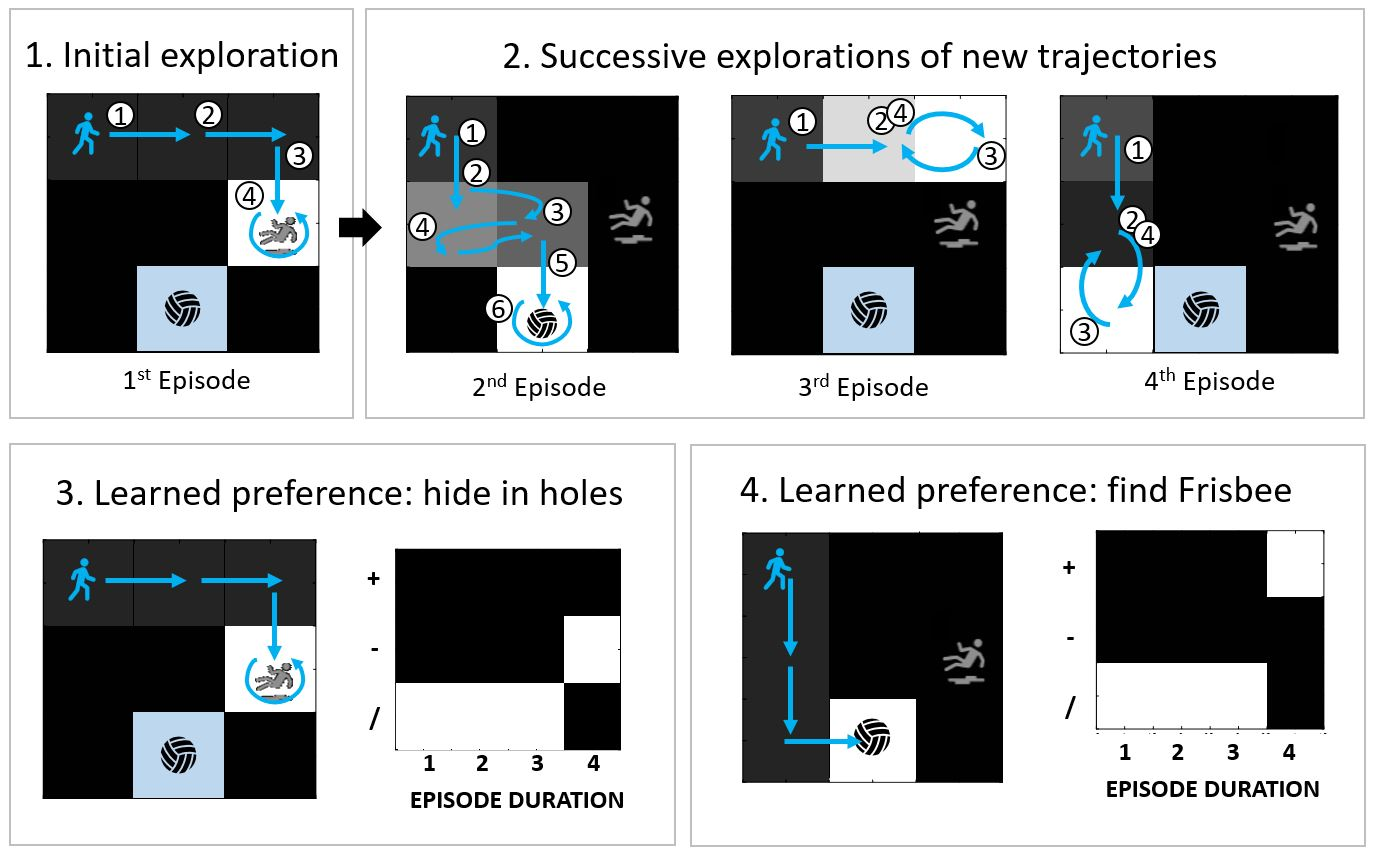}
  \caption{Parameter learning for a single reward location: results for likelihood learning presented in $1 \& 2$ and prior preference learning presented in $3 \& 4$. Blue arrows denote the trajectory taken and numbers in the circles denote the trajectory sequence. Circular arrows represent loops i.e., once in that state, the same outcome is observed till maximum number of moves reached ($15$). $5.1$ is a pictorial representation of the first episode trajectory, with no prior preference: right($1\rightarrow2$), right ($2\rightarrow3$), down($3\rightarrow6$), right($6\leftrightarrow6$). $5.2$ depicts the next $4$ episodes from the trial. $5.3$ has two figures: a pictorial representation of trajectory to hole and heat-map of the accumulated Dirichlet parameters for score ($+$is positive, --- is negative and $\ $is neutral). For this trial, there is a strict preference for holes at time step $4$. $5.4$ presents similar information, but for a goal preferring agent; pictorial representation of trajectory to goal and heat-map of the accumulated Dirichlet parameters for score. There is a strict preference for goals at time step $4$.}
  \label{fig:lll}
\end{figure}

Incrementally, we enabled learning of these parameters. First, all outcome preferences (and their Dirichlet priors) are removed. Therefore, the agent can only learn the likelihood. As a result, there is no behavioral imperative other than pure exploration \citep{Schmidhuber2006}. This set-up was simulated $15$ times and likelihood was learned in an experience dependent fashion. This results in an initial (exploratory) trajectory that covers all unchartered territory in the most efficient way possible i.e., there is no revisiting of locations that have already been encountered (Figure~\ref{fig:lll}.1). Furthermore, this behavior persists past the initial exploration, with continuous explorations via new (non-overlapping) trajectories (Figure~\ref{fig:lll}.2). This represents 'true' exploratory behavior - distinct from random action selection - of the sort only possible in a belief-based scheme. Furthermore, as there are no rewards, this behavior would be impossible to motivate from a reinforcement learning perspective, as this learning is for its own sake - not to improve reward seeking. While such an imperative could plausibly be introduced to a belief-based reinforcement learning scheme, it would have to appeal to heuristic arguments like the potential for a reward function to be introduced in the future.

Next, we equip the agent with the ability to learn outcome preferences (rather than learn about the environment). This entails updating the outcome preferences via accumulation of Dirichlet parameters, without learning the likelihood. The set-up was simulated $10$ times, for two separate kinds of outcome. During the first kind, in the absence of negative preferences, holes become attractive because they are encountered first – and this is what the agent learns about its behavior (and implicit preferences). In other words, because holes (H) are absorbing states, and the agent observes itself falling in a hole recurrently, it learns to prefer this outcome (Figure~\ref{fig:lll}.3). Similarly, in the second kind of trial, the agent finds itself recurrently acquiring the Frisbee. This causes it to exhibit preferences for acquiring Frisbee's (Figure~\ref{fig:lll}.4). These represent the capacity of active inference agents to develop into hole-seeking or Frisbee-seeking agents. As one of these outcomes becomes more familiar, the agent observes its own behavior and concludes `I am the sort of creature that enjoys spending time in holes (or with Frisbee's),' and adjusts future behavior to be consistent with this. 

This capacity is another important point of distinction with reinforcement learning approaches, where the problem is defined in terms of a pre-specified reward function. If this is the problem one hopes to solve, it is clearly undesirable for agents to develop ulterior motives. This speaks to the fundamental differences in the problems being solved by the two approaches. Under active inference, the ultimate `goal' is to maintain a coherent phenotype and persist over time. Hole-seeking agents achieve this, despite their behavior deviating from what an observer - or the designer of an AI gym game - might regard as appropriate.

\begin{figure}
  \centering
  \includegraphics[width=0.7\linewidth]{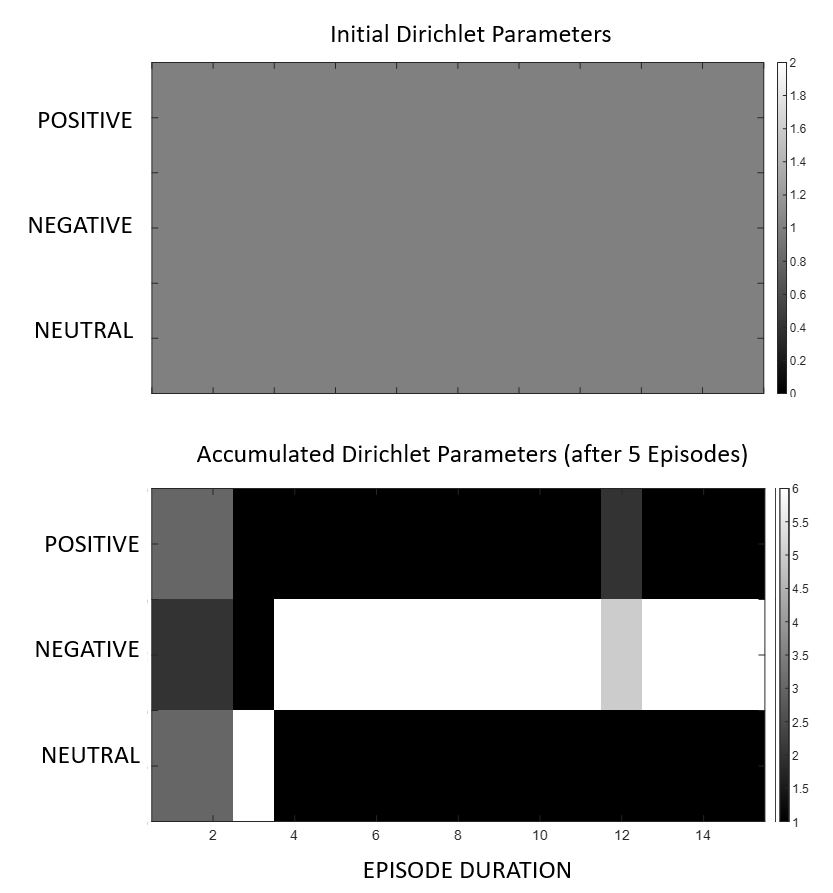}
  \caption{Learning prior outcome preferences for outcome modality, score: initial (top) and after $5$ episodes (bottom) for a single reward location}
  \label{fig:lppo}
\end{figure}

Finally, we look at the interaction between the epistemic imperatives to resolve uncertainty about the likelihood mapping and uncertainty about prior preferences. This set-up was simulated $10$ times and both likelihood distribution and prior outcome preferences learned. By parameterizing both the likelihood and prior outcome preferences with Dirichlet distributions, we induce a contribution to expected free energy that makes visiting every location attractive (i.e., every location acquires epistemic affordance or novelty). However, after a sufficient number of trials, the agent has learned (i.e., reduced its uncertainty) that it prefers to hide in holes (Figure~\ref{fig:lppo}). This causes the agent to exhibit exploitative behavior of hiding, rather than continue exploring. After $5$ trials, the agent goes straight to the hole. 

This is an interesting example of how --- by observing one’s own behavior --- preference formation contextualizes the fundamental imperative to explore.

It is important to note that the learned outcome preferences are time-dependent; i.e., the agent prefers to visit safe (F) patches for the first $3$ time points and then visit goal (G) patches with high preference (Figure~\ref{fig:lppo}). As noted, these are learned by accumulating experience (in the form of Dirichlet concentration parameters); such that uniform priors over outcomes become precise posteriors. These precise posteriors then become the agent’s preferences. Put simply, it has learned that this is the kind of creature it is.

We have observed that even in the absence of clearly defined prior preferences, active inference agents are able to learn these preferences naturally; since prior preferences are defined in terms of probability distributions, we simply define a distribution over distributions, and learn these from the data using the standard inference/gradient updates (Section \ref{subsec:optimize}). However, it is important to highlight that these learnt prior preference might be at odds with the 'reward' from the environment. This conceptualization flips rewarding states on its head, its a matter of preference not a specific scalar signal from the environment. . Concretely, we can indeed encourage AI agents to `solve’ RL environments by placing a prior preference that maximizes the observation corresponding to reward, but definitionally active inference does not require the resultant reward maximizing behavior to be considered a successful agent. As long as it can learn and then maintain a consistent set of behaviors over time -- through free energy minimization -- we consider such an agent to be successful under the active inference problem definition. 

Furthermore, by allowing various parts of the active inference framework to be learned from the environment (i.e., $\log P(o|s)$), we can infer time-dependent preferences from the environment. This is in contrast to vanilla reinforcement learning, where it is less clear how to naturally account for learning an intrinsic reward function, with many competing approaches \citep{Still2012,Mohamed2015,Deepak2017}.

\section{Discussion}
We have described active inference --- and the underlying minimization of variational and expected free energy --- using a (simplified) discrete state-space and time formulation. Throughout this review, we have suggested that active inference can be used as framework to understand how agents (biological or artificial) operate in dynamic, non-stationary environments \citep{KarlDeepTemporal2018}, via a standard gradient descent on a free energy functional. More generally, active inference can be thought of as a formal way of describing the behavior of random dynamical systems with latent states.

As noted in the formulation of active inference (see Equation~\ref{eq:v2}), epistemic foraging (or exploration) emerges naturally. This is captured by the desire to maximize the mutual information between outcomes and the hidden states on the environment. Exploration means that the agent seeks out states that afford outcomes, which minimize uncertainty about (hidden) states of affairs. In the FrozenLake simulation, this was highlighted by the initial exploratory move made by the agent, due to uncertainty about reward location. The move resolved the agent's uncertainty about the reward location and all subsequent episodes (when the reward location remained consistent) exploited this information. Note that in the formulation presented, we discussed model parameter exploration that might also be carried out by the agent --- when learning either the likelihood or prior preferences --- by having priors over the appropriate probability distributions and applying the expected free energy derivations to those parameters \citep{Schwartenbeck2019}. The simulations showed that in the absence of a reward signal from the environment, the agent could learn a niche and exhibit self-evidencing behavior. Additionally, it highlighted that due to the fundamental differences in the conceptual approach, active inference agents may exhibit Bayes--optimal behavior that is counter-intuitive from the perspective of reinforcement learning i.e., reward minimization. However, from an active inference perspective, reward is simply the sort of outcome that is preferred -– and an agent can learn to prefer other sorts of outcomes

The canonical properties presented --- with respect to decision making under uncertainty --- are usually engineered in conventional reinforcement learning schemes. However, more sophisticated formulations of reinforcement learning define a central role for uncertainty over: Q-value functions \citep{Dearden1998,Dearden2013,Osband2016,Donoghue2017}, MDP \citep{Dearden2013,Osband2016,Schulze2020} or even the reward function \citep{Sorg2012,Furnkranz2012,Haarnoja2018}. The formulation presented in \citep{Schulze2020}, incorporates uncertainty over both the model parametrisation and reward function. This suggests that there is potential to build upon (and remove components of) Bayesian reinforcement learning algorithms to render them formally equivalent to active inference. However, this may come at increased algorithmic complexity cost and loss of generalization. Additionally, these algorithmic design choices are non-trivial and may demonstrate counter-intuitive behavior \citep{Donoghue2020}. In contrast, active inference enables decision-making under uncertainty with no heuristics in play.

The simulations reveal that once the reward signal is removed the active inference exhibits information seeking behavior (to build a better model of its environment), similar to the Bayesian reinforcement-learning agent. This type of reward-free learning has been central to the curiosity literature in reinforcement learning despite by definition not being true to the definition of reinforcement learning. Concretely, these approaches induce an `intrinsic' reward using some heuristic (i.e., dynamics prediction \citep{Deepak2017}, random feature prediction \citep{Burda2018}, information gain \citep{Mohamed2015}), but this does not necessarily align with the axiomatic goal of `maximizing a numerical reward signal'; they are simply tools (i.e., inductive biases) that may lead us to achieve this, and in the case of completely absent rewards, it is unclear what the goal of reinforcement learning is (i.e., what behaviors are we ``reinforcing"?).

Our treatment emphasizes that —-- via a belief-based scheme --- active inference enables us to specify prior beliefs over preferred outcomes or not (to produce purely epistemic behavior). Practically, these can produce similar outcomes – and have behavioral equivalences to the reward function in reinforcement learning, by assigning high and low prior preferences to outcomes with positive and negative rewards, respectively. Moreover, this highlights a conceptual distinction between prior beliefs over preferred outcomes in active inference and reward functions in standard reinforcement learning. While a reward function specifies how an agent should interact with the environment, prior beliefs over preferred outcomes are a description (via some particular instantiation) of how the agent wishes to behave. Crucially, this description can be learnt over time; based on relative frequencies of outcomes encountered. This speaks to an eliminative use of Bayes optimality, which replaces the notion of reward --- as a motivator of behavior --- with prior beliefs about the outcomes an agent works towards. Conceptually speaking, this dissolves the tautology of reinforcement learning, i.e., rewards reinforce behaviors that secure rewards. Having said this, related formulations can be found in reinforcement learning: e.g., belief-based reward functions in \citep{Sorg2012,Furnkranz2012,Schulze2020}.

In active inference, agent is likely to maximize extrinsic value (c.f., expected reward) by having prior preferences about unsurprising outcomes (see Equation~\ref{eq:v1}) via the minimization of expected free energy. It is important to note that the minimization of expected free energy is achieved by choosing appropriate policies (sequences of actions). We accounted for this in the initial set-up of the FrozenLake simulation, where the agent had strong positive preference for finding the Frisbee. Additionally, hole locations were associated with strong negative preferences. In contrast, the Active inference null model with no prior preferences and no ability to learn them, encouraged exploratory behavior and the agent ended in the (G) location $44.0\%$ of the time. 

However, it is worth noting that these properties follow from the form of the underlying generative model. The challenge is to identify the appropriate generative model that best explains the generative process (or the empirical responses) of interest \citep{Sam2017}. In the FrozenLake simulation, by equipping the agents with beliefs about the current context, we were able (via the generative model and its belief updating process) to convert a learning problem into a \textit{planning as inference} problem. However, this can be treated as a learning problem by specifying a hierarchical MDP with learning capacity over the problem space. This would allow for slow moving dynamics at a higher level that account for changes in context, and fast moving dynamics at the lower level that equip the agent with the ability navigate the given instantiation of the FrozenLake \citep{KarlDeepTemporal2018}. When comparing prior preferences and rewards, we highlighted that due to no explicit prior preference for goal states, the belief-based (active inference and Bayesian RL) agents exhibit conservative behaviors; choosing to avoid the (G) state. This behavior is a caveat of the underlying generative model form --- uncertainty modeled over the location of the (G) \& (H) state --- and manipulating the prior probability distributions (or the factorization of the states) might lead to policies where agents chooses to not avoid the (G) location. 
Additionally, the generative models underlying this active inference formulation can be equipped with richer forms (e.g., via amortization) or learned via structural learning \citep{Sam2010,Sam2016}. Thus, if one was to find the appropriate generative model, active inference could be used for a variety of different problems; e.g. robotic arm movement, dyadic agents, playing Atari games, etc. We note that the task of defining the appropriate generative model (discrete or continuous) might be difficult. Thus, future work should look to incorporate implicit generative models (based on feature representation from empirical data) or shrinking hidden state-spaces, by defining transition probabilities based on likelihood (rather than latent states).

\newpage

\subsection*{Software note}
The simulations presented in this paper are available at: \href{https://github.com/ucbtns/dai}{https://github.com/ucbtns/dai}

\subsection*{Acknowledgments}
NS is funded by the Medical Research Council (Ref: MR/S502522/1). PJB is funded by the Willowgrove Studentship. KJF is funded by the Wellcome Trust (Ref: 088130/Z/09/Z). We would like to thank the anonymous reviewers for their suggestions and insightful comments on the manuscript. 

\subsection*{Disclosure statement}
The authors have no disclosures or conflict of interest.

\newpage
\bibliographystyle{unsrtnat}
\bibliography{bib}

\end{document}


\section*{Supplementary Materials}
\subsection*{Explicit parameterization of the finite horizon POMDP }\label{generativemodel}
Active inference rests on the tuple $(O,U,S,T,\Pi,R,P,Q)$ : 
\begin{itemize}
    \item A finite set of outcomes, $O$
    \item A finite set of control states or actions, $U$
    \item A finite set of hidden states, $S$
    \item $T={0,...,T}$ a finite set which determines the temporal horizon
    \item A finite set of time-sensitive policies, $\Pi$
    \item A generative process $R(\tilde{o},\tilde{s},\tilde{u})$ that generates a probabilistic outcome $o\in O$ from (hidden) state $s\in S$ and action $u\in U$
    \item A generative model $P(\tilde{o},\tilde{s},\pi,z)$  with  outcome $o\in O$, (hidden) state $s\in S$, policies $\pi \in \Pi$ and model parameters $z$.
    \item An approximate posterior $Q(\tilde{s},\pi,z)=Q(s_o|\pi)..Q(s_\uptau|\pi)Q(\pi)Q(z)$ over states $s\in S$, policies $\pi \in \Pi$ and model parameters.
\end{itemize}

The generative process describes transitions between hidden (unobserved) states in the world that generate (observed) outcomes. Their transitions depend on action, which depends on posterior beliefs about the next state. Subsequently, these beliefs are formed using a generative model of how outcomes are generated. The generative model (based on partially observable MDP) describes what the agent believes about the world, where beliefs about hidden states and policies are encoded by expectations. Here actions are part of the generative process in the world and policies are part of the generative model of the agent.

\newpage
\subsection*{Pseudo-code for active inference: belief updating and action selection}\label{pseudocode}
\begin{table}[!h]
Initialize the following: \\
Probability of seeing outcomes, given states, likelihood: $A$ \\
Probability of transitioning between states, given an action: $B$ \\
Log probability of agent's preferences about outcomes: $C$ \\
Probability of state the agent believes it is at the beginning of each trial: $D$ \\
\textbf{for} $\uptau = 1:T$ \textbf{do}
\begin{itemize}
\item[] Sample state, $s$ based on generative process
\item[] Sample outcome $o$ based on likelihood matrix $A$
\item[] Variational updates of expected states, $s$ under sequential policies\\ (gradient descent on $F$)
\item[] Evaluate expected free energy $G$ of policies $\pi$
\item[] Bayesian model averaging of expected states $s$ over policies $\pi$
\item[] Select action with the lowest expected free energy
\end{itemize}
\textbf{end} \\
Accumulation of (concentration) parameters for learning update based on learning rate  \\
\end{table}

\newpage

\subsection*{Pseudo-code for Q-Learning}\label{pseudocodeQL}
\begin{table}[!h]
Initialize the following: \\
Q-value function; $Q(s,a)$ \\
Initialize parameter for exploration; $\epsilon$\\
Specify learning rate, $\alpha$ and discount factor, $\gamma$ \\
\textbf{for} $\uptau = 1:T$ \textbf{do}
\begin{itemize}
\item[] Sample exploration rate threshold from a random uniform distribution, $U~(0,1)$
\item[] Choose action based on $max_a(Q(s,:))$ if exploration rate threshold is greater than $\epsilon$, else choose random action
\item[] Execute $a^*$ and receive $r,s'$
\item[] Update $Q(s,a):$ using $(1-\alpha)*Q(s,a) + \alpha*(r + \gamma*max_a(Q(s',:))$
\item[] $s = s'$
\item[] Update exploration parameter $\epsilon$: $\epsilon-$ \textit{decay rate}
\end{itemize}
\textbf{end} 
\end{table}

\newpage

\subsection*{Pseudo-code for Bayesian Model-Based Reinforcement Learning using Thompson Sampling}\label{pseudocodeBRL}
\begin{table}[!h]
Initialize the following: \\
$\Theta_t, \Theta_r$ as uniform \\
Probability of transitioning between states, given an action, transition model; $\Theta_t$ \\
Probability of receiving reward, given a state, reward function; $\Theta_r$  \\
\textbf{Repeat:}
\begin{itemize}
\item[] Sample $\Theta_{t,1}, ..., \Theta_{t,k} \sim Pr(\Theta_t) \forall a $
\item[] Sample $\Theta_{r,1}, ..., \Theta_{t,k} \sim Pr(\Theta_r) \forall a $
\item[] $Q^*_{\theta_{t,i},\theta_{r,i}} \Leftarrow$ solve $MDP_{\theta_{t,i},\theta_{r,i}}$
\item[] $\hat{Q}(s,a) \Leftarrow \frac{1}{k} \sum_{i=1}^{k}Q^*_{\theta_{t,i},\theta_{r,i}}(s,a)$
\item[] $a^b \Leftarrow max_a \hat{Q}(s,a)$
\item[] Execute $a^*$ and receive $r,s'$
\item[] $b(\Theta_t) \Leftarrow b(\Theta_t)Pr(s'|s,a,\theta_t)$
\item[] $b(\Theta_r) \Leftarrow b(\Theta_r)Pr(r|s,a,s',\theta_r)$
\item[] $s \Leftarrow s'$
\end{itemize}
\textbf{end} \\
\end{table}